%% file: main.tex
\documentclass[10pt,twocolumn,letterpaper]{article}
\usepackage[pagenumbers]{cvpr} 
\usepackage{graphics,graphicx,caption,float,subcaption,booktabs,xcolor,multirow,array,color,ifthen,tabu,colortbl,dblfloatfix,url,relsize,algorithm,algorithmic,amssymb,xspace,nicefrac,microtype,amsmath,amsfonts,bm}
\usepackage[pagebackref=true,breaklinks=true,colorlinks=true,bookmarks=false]{hyperref}
\usepackage[capitalize]{cleveref}
\crefname{section}{Sec.}{Secs.}
\Crefname{section}{Section}{Sections}
\Crefname{table}{Table}{Tables}
\crefname{table}{Tab.}{Tabs.}
\usepackage[english]{babel}
\usepackage{enumitem}
\usepackage{multirow}
\usepackage{capt-of,etoolbox}
\usepackage[accsupp]{axessibility}  

\def \cvprsubmission {} 
\ifx \cvprsubmission \undefined
    \newcommand{\todo}[1]{\textcolor{red}{todo: #1}}
    \newcommand{\sduggal}[1]{\textcolor{red}{#1}}
\else
	\newcommand{\todo}[1]{}
	\newcommand{\sduggal}[1]{{#1}}
\fi

\def\cvprPaperID{7777} 
\def\confName{CVPR}
\def\confYear{2022}

\makeatletter
\apptocmd\@maketitle{{\myfigure{}\par}}{}{}
\makeatother

\begin{document}

\input{tex_files/teaser}
\title{\vspace{-14mm}Topologically-Aware Deformation Fields for Single-View 3D Reconstruction\vspace{-5mm}}

\author{Shivam Duggal$\qquad$Deepak Pathak \\ Carnegie Mellon University}
\maketitle


\begin{abstract}

We present a framework for learning 3D object shapes and dense cross-object 3D correspondences from just an unaligned category-specific image collection. The 3D shapes are generated implicitly as deformations to a category-specific signed distance field and are learned in an unsupervised manner solely from unaligned image collections and their poses without any 3D supervision.
Generally, image collections on the internet contain several intra-category geometric and topological variations, for example, different chairs can have different topologies, which makes the task of joint shape and correspondence estimation much more challenging.
Because of this, prior works either focus on learning each 3D object shape individually without modeling cross-instance correspondences or perform joint shape and correspondence estimation on categories with minimal intra-category topological variations.
We overcome these restrictions by learning a topologically-aware implicit deformation field that maps a 3D point in the object space to a higher dimensional point in the category-specific canonical space.
At inference time, given a single image, we reconstruct the underlying 3D shape by first implicitly deforming each 3D point in the object space to the learned category-specific canonical space using the topologically-aware deformation field and then reconstructing the 3D shape as a canonical signed distance field.
Both canonical shape and deformation field are learned end-to-end in an inverse-graphics fashion using a learned recurrent ray marcher (SRN) as a differentiable rendering module. Our approach, dubbed TARS, achieves state-of-the-art reconstruction fidelity on several datasets: ShapeNet, Pascal3D+, CUB, and Pix3D chairs.

\vspace{-0.16in}
\end{abstract}

\input{tex_files/introduction.tex}

\input{tex_files/related_work.tex}

\input{tex_files/method.tex}
\input{tex_files/experiments.tex}
\input{tex_files/conclusion.tex}

\section{Acknowledgement}
We would like to thank Shamit Lal, Jason Zhang, Alex Li,  Ananye Agarwal for feedback on the paper and Chen-Hsuan Lin for providing details on the SoftRas baseline and the Pascal3D dataset. We are grateful to Ankit Ramchandani for help before the deadline by painting chair meshes for texture transfer experiment. This work is supported by DARPA Machine Common Sense program.

{\small
\bibliographystyle{ieee_fullname}
\bibliography{egbib}
}

\newpage
\appendix
\input{tex_files/supp.tex}

\end{document}

%% file: tex_files/teaser.tex

\newcommand\myfigure{%
\vspace{-0.1in}
\centering
\includegraphics[width=0.95\linewidth]{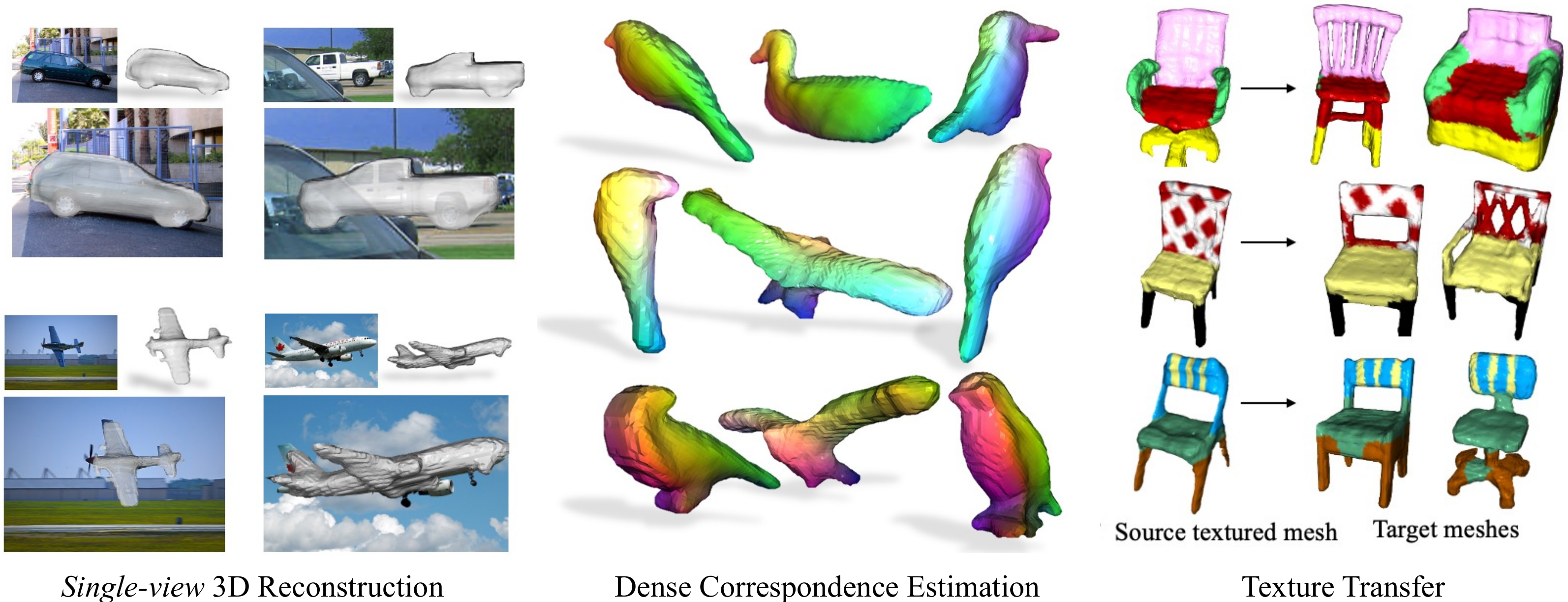}
\vspace{-0.07in}
\captionof{figure}{Given an unpaired image collection \sduggal{(with camera poses)} of an object category at training time, our approach learns to: (a) {\textit{reconstruct the underlying 3D}} given only a single image at test time, and (b) \textit{{model dense 3D correspondences}} across category instances. 
The learned correspondence field is \sduggal{articulation-aware}, topologically-aware and inherently captures structural properties of the category, enabling the task of dense \textit{{texture transfer}}. Videos and code at~\url{https://shivamduggal4.github.io/tars-3D/}}
\vspace{0.15in}
\label{fig:teaser}
}

%% file: tex_files/introduction.tex
\section{Introduction}
Learning to understand the 3D geometric world underlying our 2D observation snapshots has been a longstanding problem in computer vision, yet the generalization is nowhere close to that in learning to recognize 2D concepts \cite{mask-rcnn, He2015, dosovitskiy2020vit}. The reason is rather unsurprising: the lack of scalable ways to obtain 3D supervision in the wild, be it multiple views of the same object or GT shape.
Unlike the current visual systems, humans can infer 3D structure just from a single image (even under large occlusions). 
If our (deep) learning models
have to \sduggal{develop such a capability,}
we must first figure out how to
understand the
3D structures from just an unaligned and diverse 2D image collection -- the kind of data available \emph{in abundance} on the web. However, any such approach must answer a fundamental question first -- how should one represent the 3D structure?

\begin{figure*}[t]
\centering
\includegraphics[width=0.92\linewidth]{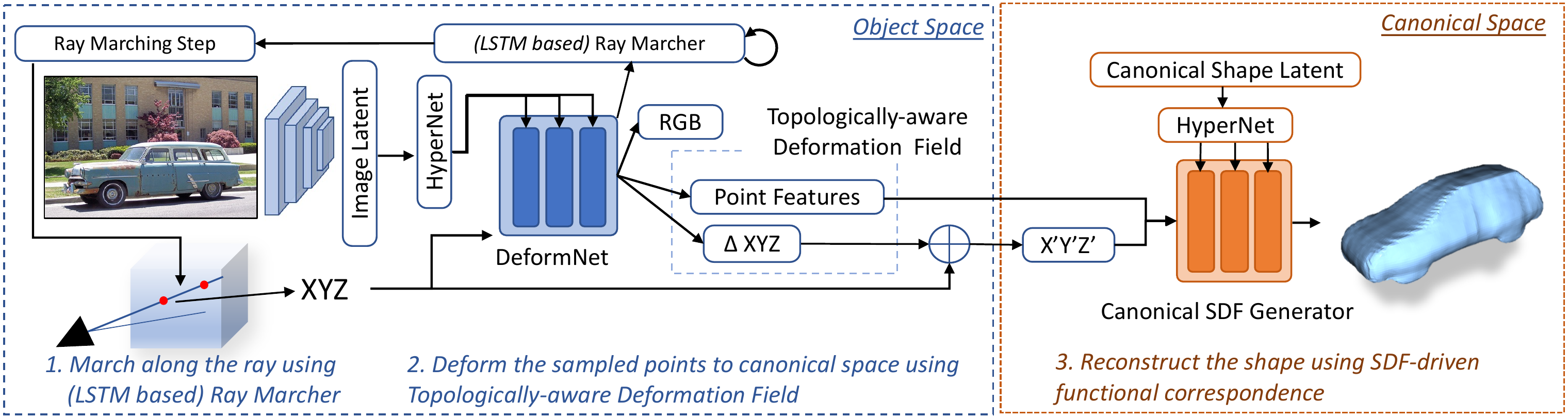}
\vspace{-2mm}
\caption{Overview of TARS: Given a single image, we first map a 3D point in object space to a higher-dimensional canonical space using our learned topologically-aware deformation field. The canonical point is then mapped to its SDF value using the Canonical Shape Generator module. We leverage an LSTM-based differentiable renderer to guide the learning of deformation and signed distance fields.}
\label{fig:pipeline}
\vspace{-5mm}
\end{figure*}


Looking at the research in recent years, there is an overwhelming evidence in support of implicit representations credited to the advancement in neural implicit modeling~\cite{mildenhall2020nerf, niemeyer2020differentiable, ONet, park2019deepsdf, PiFU, chen2018implicit_decoder, dist}. While these implicit representations have attained the gold standards of high-fidelity reconstruction, they still rely on either 3D GT shape or dense multi-view supervision not only during training but sometimes also at inference~\cite{mildenhall2020nerf}, making them difficult to apply to the internet of images. Recent works~\cite{yu2020pixelnerf, zhang2021ners, jain2021putting,SRF} have attempted to cut down the requirement of multi-view images from $100$s to $2-10$. However, as long as any method needs more than a single image, it can not be used to 3D-fy trillions of images on internet -- the setting considered in this paper. 
What kind of signals can we exploit from 2D image collection of a category at training time, that can help generate 3D for an unseen 2D image 
at test time? We turn to Plato.

\begin{figure*}[t]
\centering
\includegraphics[width=0.80\linewidth]{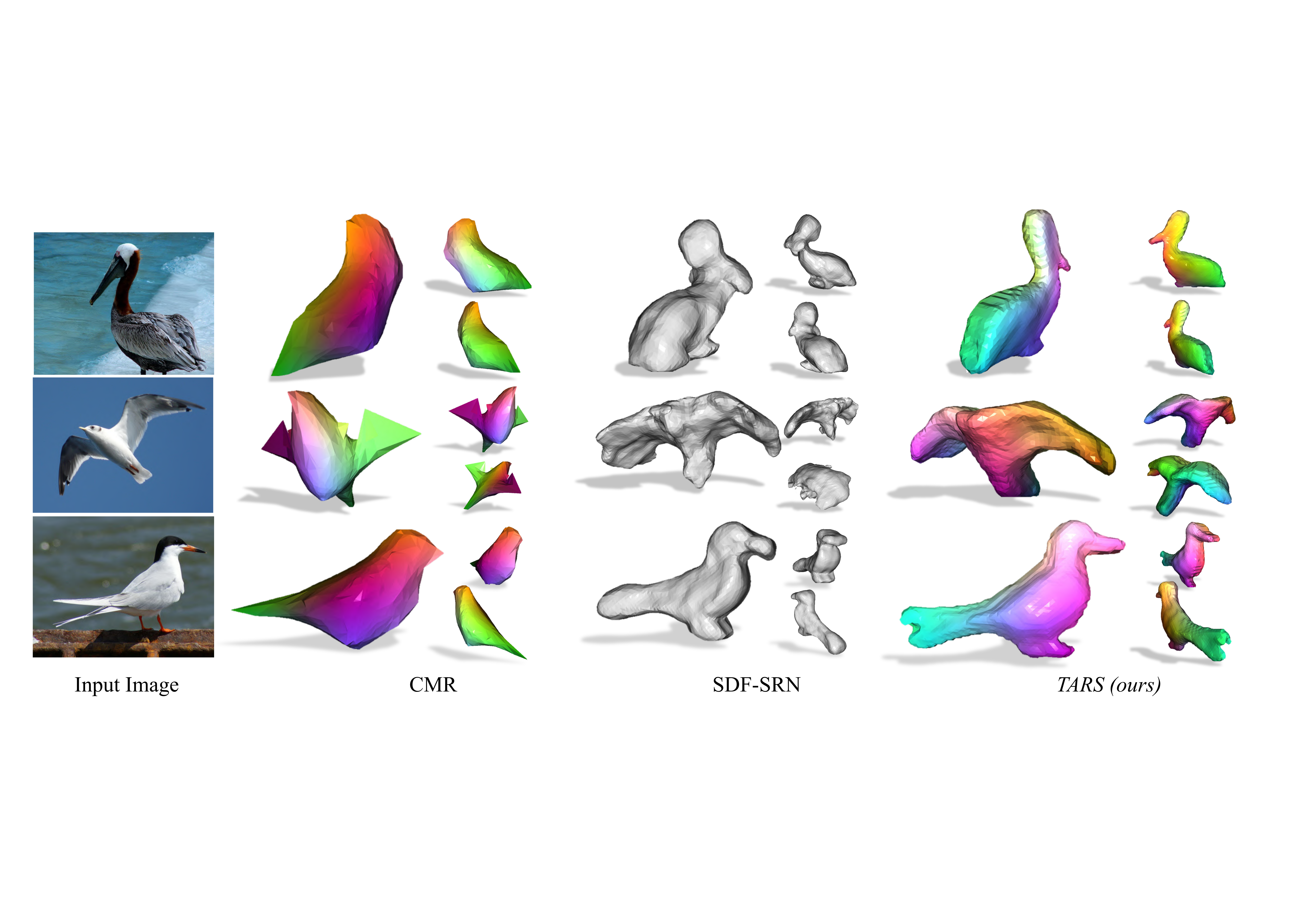}
\vspace{-2mm}
\caption{3D Reconstruction on CUB-200-2011. Compared to prior works, not only our reconstructions are of higher fidelity, but the learned (color-coded) deformation fields are also articulation aware (eg: rotated heads, open wings). Unlike CMR, we do not hard-code symmetry.}
\label{fig:cubs_comparison}
\vspace{-5mm}
\end{figure*}

Plato's philosophy of ``Theory of Forms'' relates every object in reality to a particular form or an idea (a platonic ideal). His famous example of ``cupness'' says that while there exists many cups, there is only one ``idea'' of cupness. We believe this is closely tied to human perception of objects. For instance, when we play the game of pictionary \cite{QuickDraw}, given just a category-level description of an object, we can generally draw its high-level (category-level) representation.
Only when we are provided with more observations or properties 
of an object (eg: a chair ``with arms'', ``an SUV'' car, an airplane ``with wider wings''), we are able to draw that specific instance of the category.
This philosophy has been classically adopted in deformable models~\cite{3DMM} but require 3D supervision. More recently, 
with
the advent 
of differentiable renderers \cite{liu2019softras, kato2018renderer, chen2019dibrender}, this has been adopted for estimating 3D from a single image~\cite{cmrKanazawa18, umr2020, ucmrGoel20, bhattad2021view}.
However, these methods are restricted to categories with minimal to no intra-category topological variations because of fixed mesh connectivity and absolute reconstructions are also of lower fidelity compared to implicit methods \sduggal{(see CMR results in Figure~\ref{fig:pascal_comparison_chairs})}.

3D objects that belong to the same (``platonic'') category generally inherit similar structural and semantic properties.
In this work, we follow this ideology and propose a 3D reconstruction algorithm, which can: (a) learn from just an unaligned 2D image collection without any 3D or multi-view supervision at training and inference; (b) generalize to topologically diverse categories like chairs which mesh-based approaches can't; and (c) can learn dense 3D correspondence across different instance shapes for free by mapping the object instances to the category mean, allowing the model to exploit cross-image similarity.
These intra-category correspondences are very beneficial for numerous vision and graphics tasks: geometry/shape understanding \cite{nrsfm_kanade, wang2020deep, SMPL:2015, SMAL}, 3D manipulation \cite{3DMM, SMPL:2015, SMAL}, 2D image synthesis \cite{SRF,Face2Face, weng2018photo}, 2.5D depth estimation \cite{schoenberger2016mvs,Luo-VideoDepth-2020,zbontar2016stereo}, etc.

However, simply extending implicit models and learning implicit dense correspondences between topologically varying objects with just single view supervision is not straight-forward. This is because of inherent continuous nature of MLPs used by implicit shape modeling techniques and the inherent discontinuities in correspondence field between any two topologically different objects. 
For any two instances with different topologies, correspondence field has to be dis-continuous in order to map one structure to the another. Please refer to supp. section \ref{sec:topological_understanding} for more understanding. To overcome this issue of implicitly learned deformation fields, we propose \emph{topologically-aware deformation fields}.

Given an object image, we first map a 3D point in the object space to the corresponding 3D-point in the category-level canonical space using our \emph{DeformNet module}. Then, to address the above issue of implicit deformation fields and to learn correspondences between topologically varying shapes, we take inspiration from Level Set Method (LSM) \cite{Osher01levelset, Osher88frontspropagating}. Level Set Methods support topological merging/breaking of shapes by representing any shape as a zero-level crossing of a higher-dimensional function. Inspired from them, \emph{we transform our 3D canonical points to a higher-dimension by concatenating them with learned object-space point features.} We then estimate the underlying shape by mapping the higher-dimensional canonical points to the corresponding SDF values using the \emph{Canonical Shape Generator} module. A high-level overview of our approach is shown in Figure~\ref{fig:pipeline}. 

We dub our approach \textbf{TARS} (\textbf{T}opologically-\textbf{A}ware \textbf{R}econstruction from \textbf{S}ingle-view), see overview in Figure~\ref{fig:pipeline}.
We utilize a differentiable renderer in our pipeline to learn both the deformation and the shape reconstruction modules using image collections containing single-view RGB observation, corresponding GT camera pose and object silhouette. Our differentiable render (inspired from SRN\cite{SRN}) is a form of a neural render \cite{neural-rendering} which takes in features of a 3D point in the object space (visible from the input viewpoint) and predicts its corresponding depth value as seen from the input view point. Thus, during training we learn the object shape in two ways: (a) 2.5D depth representation learned using object-level features (via differentiable renderer), (b) 3D SDF learned using canonical shape features (via canonical shape generator). By enforcing the consistency between the two shape representations, we are able to effectively learn the correspondence field. Since this shape consistency is the courtesy of the differentiable renderer, we term it as the \emph{differentiable render consistency} in the following sections.

The closest approach to ours is SDF-SRN \cite{SDF-SRN}, a neural implicit shape modeling approach for single-view 3D reconstruction. Unlike us, they directly map a point in the object space to the corresponding SDF value and hence do not output dense correspondences across object instances. 

\sduggal{We validate the effectiveness of our learned shapes on multiple datasets: ShapeNet \cite{Chang2015ShapeNetAI}, Pascal3D+ \cite{Pascal3D}, CUB-200-2011 \cite{WelinderEtal2010} and Pix3D chairs \cite{pix3d}. Our method, TARS, outperforms priors works in term of 3D reconstruction fidelity and generates shapes with better global structure and finer instance-specific details. Unlike prior deformable single-view reconstruction works \cite{cmrKanazawa18, ucmrGoel20, umr2020}, which are restricted to categories like cars/ cubs, we take the first major step in modeling topologically-challenging categories (chairs). The learned topologically-aware deformation field captures structural properties of the category (without any supervision), thus enabling dense texture-transfer (Figure \ref{fig:teaser}).}



%% file: tex_files/related_work.tex

\section{Related Work}

\begin{figure*}[t]
\centering
\includegraphics[width=0.82\linewidth]{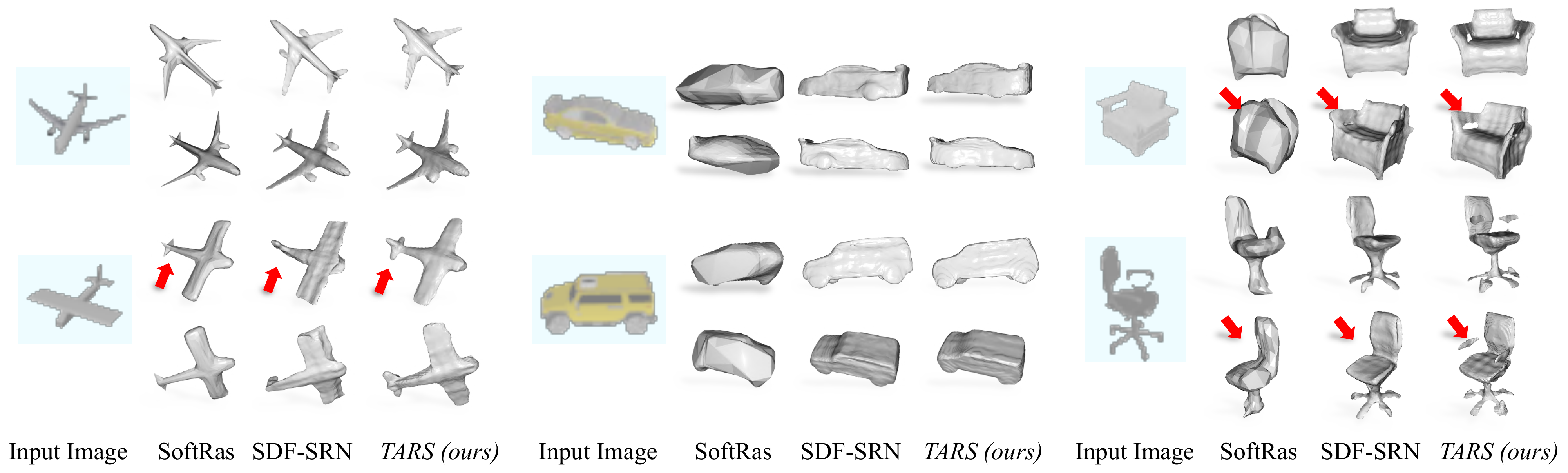}
\vspace{-2mm}
\caption{3D Reconstruction on Shapenet. Compared to mesh-based SoftRas, both neural implicit approaches yield higher 3D fidelity. Our approach additionally provides correspondences (even across topologically-varying structures), while matching SDF-SRN's shape fidelity.}
\label{fig:shapenet_comparison}
\vspace{-5mm}
\end{figure*}

Reconstructing 3D from 2D observations has been an actively studied problem \cite{schoenberger2016mvs,Newcombe2011KinectFusionRD,shape_from_silhouette,park2019deepsdf,ONet,chen2018implicit_decoder}. 
Until recently, the majority of the high-fidelity reconstruction results were credited to the availability of some form of 3D data \cite{qixing_huang_retrieval,Newcombe2011KinectFusionRD}, and because of this, the majority of the success had been restricted to synthetic datasets \cite{Chang2015ShapeNetAI}. Reconstruction of real-world 3D shapes was either done by transferring the learned synthetic models to real-world objects \cite{Bechtold2021HPN,sdf_secrets,marrnet,shapehd} or required special 3D sensors \cite{sdflabel,sdf_secrets,Newcombe2011KinectFusionRD}. However, collecting dense 3D data is cumbersome, challenging, and even not possible for certain categories (eg: birds). With advancements in inverse graphics and differentiable rendering \cite{Loper2014OpenDRAA,kato2018renderer,liu2019softras, liu2020dist, drcTulsiani17,chen2019dibrender}, the requirement of 3D supervision has been significantly reduced. More recently, significant progress has been made in this direction 
and the reconstruction quality has reached its golden standard, particularly thanks to the combination of neural implicit representations and differentiable rendering \cite{mildenhall2020nerf, SRN, DVR, yariv2020multiview}. However, the majority of these works still require dense supervision in form of paired multi-view images. Such a setting may not be possible for the internet of images. Our work, TARS, focuses on further mitigating the dependency on dense supervision by operating only on single-view data.

\vspace{-5mm}
\paragraph{3D Reconstruction with Single-view Supervision:}
The task of single-view 3D reconstruction has been comparatively less-explored. Kar \etal \cite{Kar2015CategoryspecificOR}, Kanazawa \etal \cite{cmrKanazawa18} took a major step in this direction by learning 3D structures from a large collection of unpaired images. They learned to reconstruct the underlying shape by learning the deformations on top of a (learned) category-specific mean mesh. Further research along this direction focused on reducing supervision \cite{ucmrGoel20, umr2020}, enhancing geometry \cite{implciit_mesh_shubham, chen2019dibrender} and texture fidelity \cite{bhattad2021view}. However, these works are restricted to the reconstruction of object categories with topologically similar instances (eg: birds). Like CMR  \cite{cmrKanazawa18}, we leverage the structural knowledge embedded in image collection in form of learned deformations to a learned category-specific mean shape but overcome their topological restrictions. Recently, Lin \etal \cite{SDF-SRN} directed the success of neural implicit modeling \cite{SRN} to the task of single-view 3D reconstruction and achieved state of the art results in terms of geometric fidelity. Our work further boosts the fidelity standards by jointly learning category-specific deformations and SDF fields.

\vspace{-4mm}
\paragraph{Neural Rendering:}
Recent works \cite{mildenhall2020nerf, wang2021neus, DVR, yariv2020multiview,Oechsle2021ICCV,dist, SRN} for rendering implicit surfaces have majorly leveraged some form of ray-tracing (ray marching, volumetric  or surface rendering). The recent survey on neural rendering \cite{Tewari2020NeuralSTAR} classifies the rendering as: (a) \emph{image-based rendering approaches}, which generate 2D content without explicitly modeling 3D (by transforming/ warping the input images) (b) \emph{explicit 3D based approaches}. In our work, we utilize SRN \cite{SRN} as our neural renderer. SRN \cite{SRN} performs LSTM-based ray marching to implicitly generate a 2.5D depth map corresponding to the input image. Therefore, SRN lies at the intersection of image-based and explicit rendering approaches. By using SRN \cite{SRN} as image-based renderer, we learn shapes in two ways: image to depth map translation learned using object-space features, and image to SDF learned using canonical-space features. Consistency between the two shape representations is the key contributor to our performance.

\vspace{-5mm}
\paragraph{3D Reconstruction with Dense Correspondences:}
Learning category-specific deformable shapes have been found to be prominently useful for 3D reconstruction \cite{3DMM, SMPL:2015, Kar2015CategoryspecificOR, cmrKanazawa18,Engelmann2017SAMPSA}.  These approaches generally learn instance shapes as deformations to the initial shape bases. Prior works along this line (reconstruction via deformation) have reconstructed 3D shapes either in volumetric grid representation \cite{Engelmann2017SAMPSA,Wang2019DirectShapePA} or mesh representation \cite{SMPL, Kar2015CategoryspecificOR, cmrKanazawa18}. We learn both the deformation field and the 3D shape (signed distance field) implicitly. Unlike deformations to mesh, learning deformations to an implicit field is much more challenging, because of the loss of explicit structure (mesh connectivity). Recently, \cite{deng2021deformed, zheng2020dit} learned category specific deformation and signed distance fields implicitly. However, unlike our approach (TARS), they require dense 3D supervision during training.

%% file: tex_files/method.tex
\section{Method}


Given a single image of an object, our goal is to reconstruct the underlying 3D shape. Rather than directly reconstructing the shape, we learn to reconstruct the object's 3D shape by implicitly mapping it to a (learned) category-specific canonical shape. In order to do so, we leverage a category-specific collection of unpaired object images (along with camera poses and object silhouettes) as our training corpus. This allows us to incorporate category-specific knowledge into our shape reconstruction pipeline. 
Our pipeline (as shown in Figure~\ref{fig:pipeline}) consists of three core components: (a) \emph{Deformnet}, for prediction of topologically-aware deformation fields, (b) \emph{Canonical Shape Generator}, for reconstruction of object's 3D shape (as SDF) and, (c) \emph{Differentiable Renderer module}, to render the learned SDF and hence guide the learning of Deformnet and Canonical Shape Generator during the training phase. In the following sections, we first discuss these modules and then stick them together to define our inference and training regimes.

\begin{figure*}[t]
\centering
\includegraphics[width=0.95\linewidth]{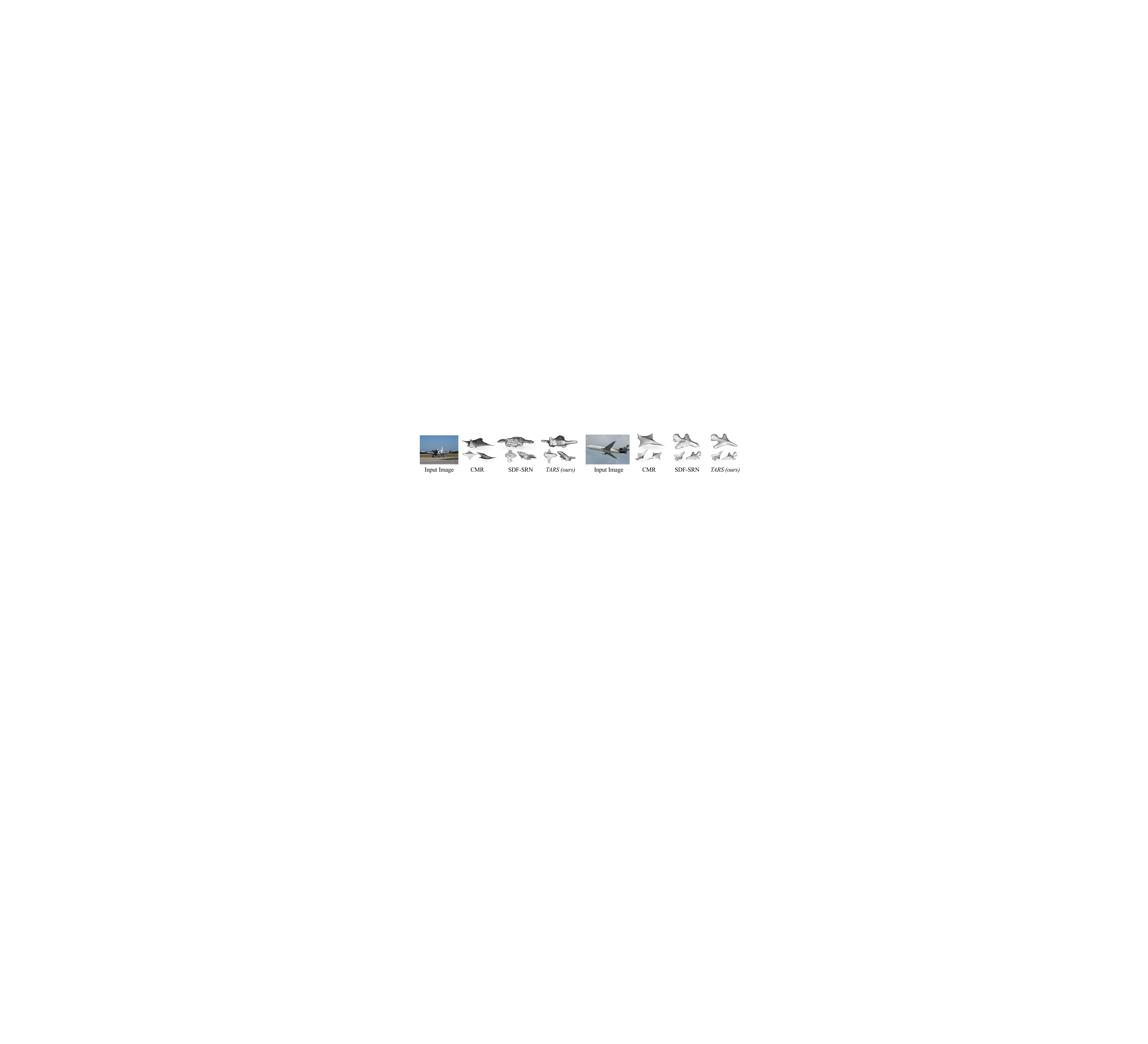}
\vspace{-2mm}
\caption{3D Reconstruction on Pascal3D+ planes. Compared to prior works, our approach performs well even with the challenging real-world observations, generating 3D shapes which are less noisy and better represent the overall structure of the ground-truth shapes.}
\label{fig:pascal_comparison_planes}
\vspace{-5mm}
\end{figure*}

\subsection{Topologically-Aware Deformation Fields}
\paragraph{Learning Implicit Deformation Fields:}
The goal of DeformNet ($g$) is to learn dense 3D point deformations from object-space to canonical-space. More formally, given an image $I$, and a 3D point ($x_\mathrm{{object}}$) in object space, the deformation estimation task is defined as:
\vspace{-2mm}
\begin{align}
   x_\mathrm{object} + g(x_\mathrm{{object}}, I) = x_\mathrm{{canonical:3D}}
    \label{eq:1}
\end{align}
where $x_\mathrm{{canonical:3D}}$ is the corresponding point in the canonical space. The mapping between the two points ($x_\mathrm{{object}}$ and $x_\mathrm{{canonical:3D}}$) is learned by leveraging signed distance function (SDF) as the functional map \cite{Ovsjanikov_functionalmaps:} between the two spaces i.e. SDF of $x_\mathrm{{object}}$ w.r.t object's surface should be same as SDF of $x_\mathrm{{canonical:3D}}$ w.r.t canonical shape surface.

We implement DeformNet module as an MLP. To learn the deformation field, we condition the DeformNet module on the input image through a hyper-network. The input image is first passed through ImageNet pre-trained ResNet encoder \cite{He2015} to generate a latent-code. Inspired from \cite{SRN, sitzmann2019siren, SDF-SRN, MetaSDF}, the computed latent code is then used by the hyper-network to predict the weights for the DeformNet MLP. We observe that using the hyper-network rather than directly learning the weights of the MLP leads to smoother shapes.

\begin{figure*}
\centering
\includegraphics[width=0.75\linewidth]{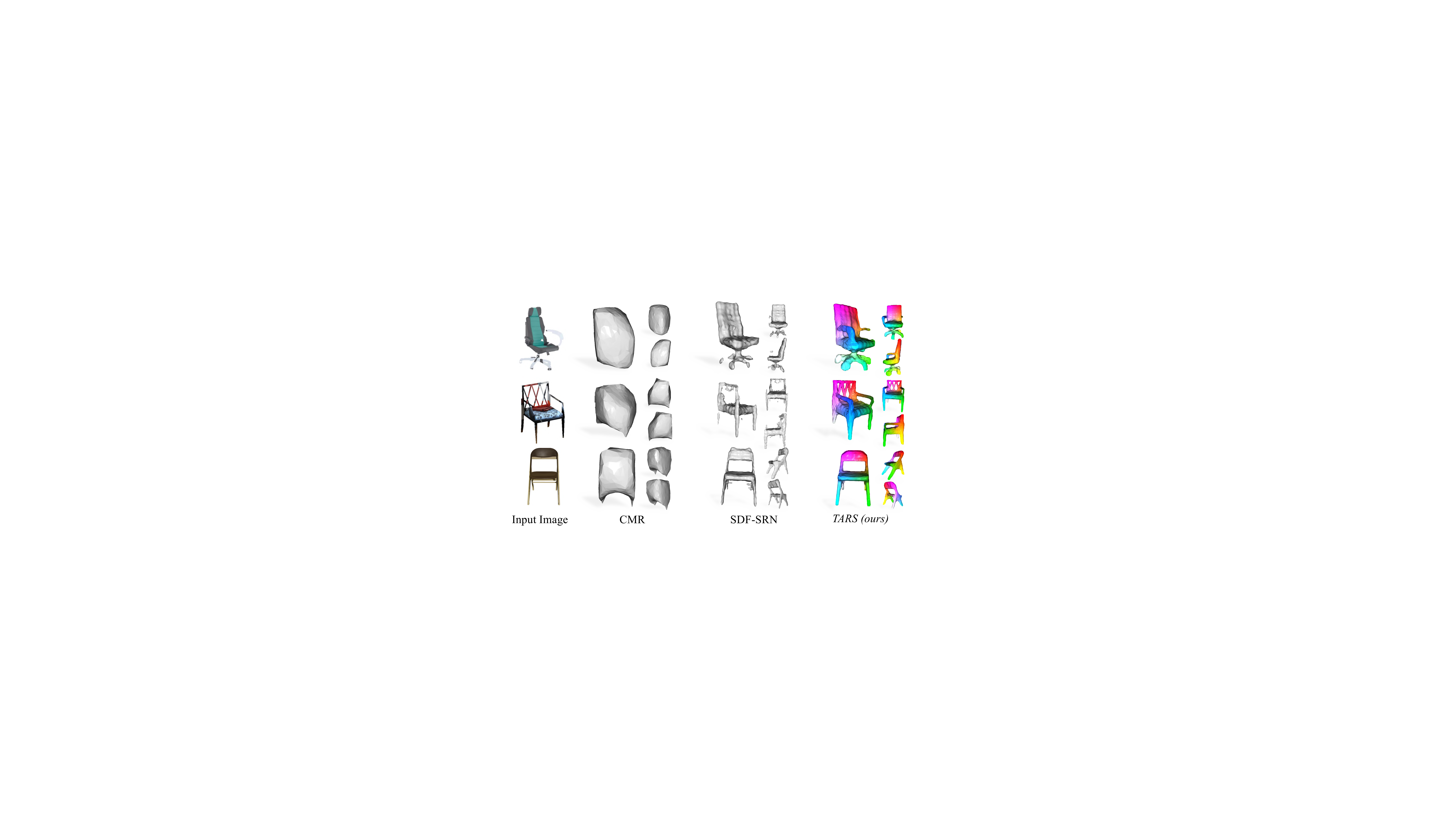}
\vspace{-2mm}
\caption{3D Reconstruction on Pascal3D+ (default) Chairs. Compared to the implicit approaches, CMR completely fails to model the topologically-varying chairs category. (Color denotes mapping to canonical space)}
\label{fig:pascal_comparison_chairs}
\vspace{-5mm}
\end{figure*}

\vspace{-4mm}
\paragraph{Point-features for Learning Topologically-Aware Deformation Fields:} Unlike prior works \cite{cmrKanazawa18, ucmrGoel20, umr2020, bhattad2021view}, our goal is to reconstruct 3D shapes even for object categories with large intra-category topological variations (eg: see chairs in Figure~\ref{fig:teaser}, Figure~\ref{fig:pascal_comparison_chairs} and Figure~\ref{fig:pix3d_shapenet}). In order to do so, we need to ensure that our deformation field can map any input object with an arbitrary topology to the canonical shape with a fixed topology. However, learning such a deformation field using an MLP is a challenging task. This is because of the continuous nature of the MLP. While the continuous nature of MLP 
\sduggal{assists in learning the 3D shapes implicitly,}
such a property hurts the learning of cross-object deformations. This is because the deformation field between objects of different topologies could be discontinuous (supp. Figure~\ref{fig:supp_topological_deformation_example}). To overcome this issue and effectively learn both the deformation and the shape fields, we take inspiration from the level-set methods (LSM) theory. LSM \cite{Osher01levelset} allow topological merging and breaking of structures by modeling any surface as a zero-level crossing of a higher-dimensional function. We take inspiration from these works \cite{Osher88frontspropagating, 10.1007/BFb0086904, Osher01levelset, homayounfar2020levelset}, and learn a higher-dimension deformation field (7D in our implementation) instead of previously learned 3D deformation fields. Concurrently, Park \etal \cite{park2021hypernerf} proposed similar insights for learning deformations between multiple views of the same object instance. To learn the higher-dimensional deformation field, we also learn object-space point features, $h(x_\mathrm{{object}})$ using the intermediate-level features of DeformNet, alongside learning the above-defined 3D deformation field (Eq.~\ref{eq:1}). Thus, we deform a point ($x_\mathrm{{object}} \in \mathbb{R}^3$) in object space to a higher-dimensional canonical point ($x_\mathrm{{canonical:HD}} \in \mathbb{R}^{3+k}$) ($k$ equals the dimension of the learned point features), where $x_\mathrm{{canonical:HD}}$ is simply the concatenation of 3D canonical point ($x_\mathrm{{canonical:3D}}$) and learned point features, $h(x_\mathrm{{object}})$. We notice that learning these point features leads to reconstructions with sharper details and better preservation of topology of GT shape (see Figure~\ref{fig:ablation_deformation_field}).


We also predict view-independent RGB value of the input 3D point using intermediate-level features of DeformNet.

\subsection{Canonical Shape Reconstruction} Now that we have deformed the 3D points in object space to the corresponding points in canonical space, our next task is to learn the 3D shape in form of SDF field. To estimate the SDF value of the 3D object point ($x_\mathrm{{object}}$), we pass the corresponding higher-dimensional canonical point ($x_\mathrm{{canonical:HD}}$) through the Canonical Shape Generator module ($f$). We learn the weights of shape generator using a hyper-network. The hyper-network is conditioned on a canonical shape latent-code ($L$), which is jointly learned during training. The canonical reconstruction task is defined:
\vspace{-3mm}
\[f(x_\mathrm{{canonical:HD}}, L) = s \vspace{-2mm}\]
, where $s$ is the signed-distance value of $x_\mathrm{{canonical:3D}}$ w.r.t the canonical shape surface (also equals signed distance value of $x_\mathrm{{object}}$ w.r.t input object's surface by the property of the established functional map).

\subsection{Differentiable Renderer Consistency}
In this section, we define the differentiable renderer and our proposed differentiable renderer consistency term which are used in our training pipeline (Figure~\ref{fig:pipeline}). The differentiable renderer is used to generate 2D renderings of the learned 3D shape during training, which are then compared against input object's GT 2D observations (RGB map and silhouette). Following \cite{SDF-SRN}, we utilize SRN\cite{SRN} as our LSTM-based differentiable renderer. The renderer works by performing the ray marching procedure, where every marching step is learned in form of a depth estimate from the current 3D point along the current camera ray direction. Please refer to SRN paper\cite{SRN} for more details.

Prior works on deformation-driven inverse graphics \cite{cmrKanazawa18,ucmrGoel20,park2021nerfies,tretschk2021nonrigid} rendered the learned 3D shape (be it mesh, density field or signed distance field) to compute the loss terms for training. Unlike them, instead of rendering the signed-distance field (like \cite{wang2021neus, dist}) learned by the Canonical Shape Generator, we utilize SRN as an image-based neural renderer. It takes as inputs the intermediate-level object features of the DeformNet module and predicts a 2.5D depth map of the input object (as viewed from input viewpoint). \emph{This allows us to enforce consistency between the two shape representations learned in our training pipeline: (a) 2.5D depth map learned via object-space point features, (b) 3D signed distance value learned via canonical-space point features.} We enforce the signed distance value of the last and the second last 3D points along the (renderer's) marched rays (for rays hitting the object) to be $\mathrm{-ve}$ and $\mathrm{+ve}$ respectively. The consistency term has been adopted from SDF-SRN\cite{SDF-SRN}. However, they established this consistency between the two shape representations learned in the same object space. Unlike them, our purpose to utilize such a consistency to allow efficient learning of the deformation field.

\subsection{Inference and Training regimes}
\paragraph{Inference:} In order to reconstruct the 3D shape underlying the input image, we first densely sample points within a unit-cube and map them from object space to canonical-space using the topologically-aware deformation field. The SDF values of the deformed object points are then estimated using Canonical Shape Generator. Finally, we utilize marching cubes \cite{MarchingCubes} to generate a 3D mesh from the learned SDF field. 

\vspace{-4mm}
\paragraph{Training:} Our training procedure is similar to the recent image-based implicit shape modeling and novel-view synthesis works \cite{mildenhall2020nerf, SRN, SDF-SRN}. We begin with shooting variable number of camera rays from the input camera viewpoint. \sduggal{We iteratively march along each camera ray and for each 3D point ($x_\mathrm{{object}}^i$) along the ray, we predict: (a) corresponding canonical point ($x_\mathrm{{canonical:HD}}^i$) using the DeformNet, (b) corresponding SDF value of the canonical point using the shape generator and, (c) the ray marching step ($d^i$) using the LSTM renderer.}
The next 3D point along the ray is then estimated as: $x_\mathrm{{object}}^{i+1}$ = $x_\mathrm{{object}}^{i} + d^i \overrightarrow{r}$ ($\overrightarrow{r}$ is the unit ray direction). The above procedure is repeated $\mathrm{n}$ times $(i \in \mathrm{n})$ along all rays (where $\mathrm{n}=$ \# of ray marching steps). Our training objective is similar to SDF-SRN's \cite{SDF-SRN} and is defined as:
\vspace{-3mm}
\begin{align*}
	\ell_{\textrm{total}} = \ell_{\textrm{rgb}} \;+\; \ell_{\textrm{sdf}} \;+\; \ell_{\textrm{reg}}
\end{align*}

\textbf{RGB loss term} ($\ell_{\textrm{rgb}}$) is simply the mean-squared error between a 3D point's predicted RGB value and the GT pixel intensity of the corresponding rendered pixel. \\

\vspace{-3mm}
\textbf{SDF loss term} ($\ell_{\textrm{sdf}}$) enforces the proposed differentiable renderer consistency. For camera rays intersecting the 3D object (guided by the GT object silhouette), SDF loss term enforces all points other than the last ray point to have SDF $>$ 0 (outside the object surface) and the last ray point to have SDF $<$ 0 (inside the surface). SDF value is penalized to be greater than 0, for all points on non-intersecting rays. Following SDF-SRN \cite{SDF-SRN}, we also utilize the distance transform of the input object mask to penalize the lower-bound of the SDF values of points lying outside the surface. Please check SDF-SRN \cite{SDF-SRN} for more details on the distance-transform loss term. \\

\vspace{-3mm}
\textbf{Regularization terms} ($\ell_{\textrm{reg}}$):\textbf{} We utilize two regularization terms: Eikonal loss ($\ell_{\textrm{eik}}$) and Deformation smoothness ($\ell_{\textrm{def}}$).
We apply eikonal loss on canonical points ($x_\mathrm{{canonical:3D}}$) and def. smoothness on object-space points.
\vspace{-4mm}
\begin{align*}
	\ell_{\textrm{eik}} = \sum_{x \in \Omega} ||\nabla f(x + g(x, I)) - 1||_{2}^{2}
\end{align*}
\vspace{-7mm}
\begin{align*}
	\ell_{\textrm{def}} = \sum_{x \in \Omega} ||\nabla g_x(x) + \nabla g_y(x) + \nabla g_z(x)||_{2}^{2}
\end{align*}
For both the regularization terms, we sample from the unit cube ($\Omega$) bounding a normalized 3D object.
	


%% file: tex_files/experiments.tex

\section{Experiments Details}

\paragraph{Datasets}
We train and evaluate our proposed approach as well as the baselines on \sduggal{following datasets: Shapenet \cite{Chang2015ShapeNetAI}, Pascal3D+ \cite{Pascal3D}, CUB-200-2011 \cite{WelinderEtal2010} and Pix3D chairs \cite{pix3d}.} Each training example consists of cropped RGB image (centered around the object), corresponding segmentation map and camera pose.
At inference time, we only need the object image as input. 
Please check supp. for more details.

\begin{figure}[t]
\centering
\includegraphics[width=0.86\linewidth]{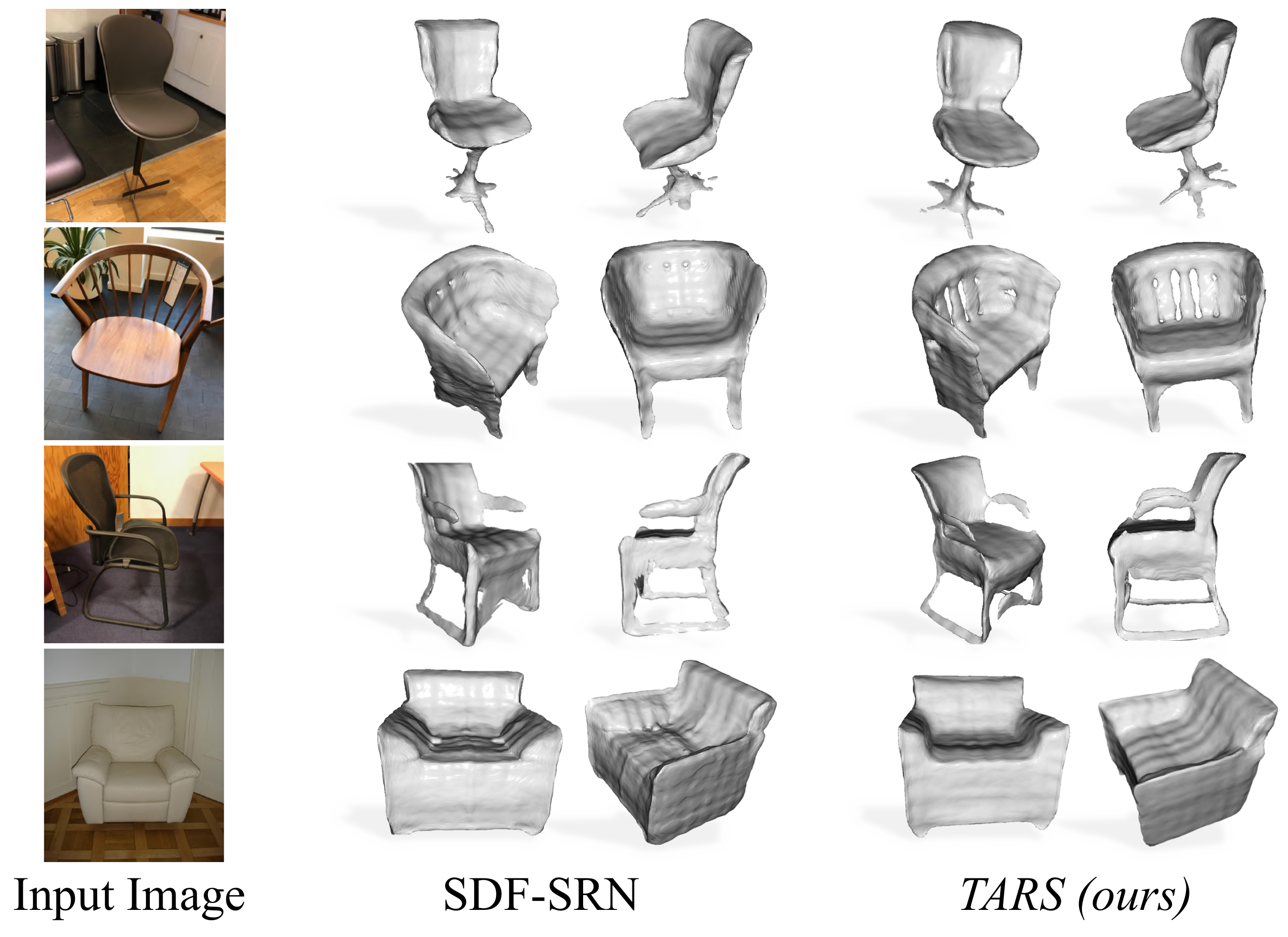}
\vspace{-3mm}
\caption{\small 3D Reconstruction on Pix3D (trained on Shapenet).}
\label{fig:pix3d_shapenet}
\vspace{-6mm}
\end{figure}

\vspace{-4mm}
\paragraph{Baselines}
We compare against the state-of-the-art methods on the task of 3D reconstruction: (a) SoftRas \cite{liu2019softras}: rasterization-based differentiable mesh renderer. (b) SDF-SRN \cite{SDF-SRN}: neural implicit modeling approach for single-view reconstruction. It is closest to ours but does \textit{not} learn any correspondences across instances. (c) CMR \cite{cmrKanazawa18}: deformation-driven mesh reconstruction approach which uses NMR \cite{kato2018renderer} as the differentiable renderer and also learns dense correspondences. We achieve state-of-the-art quantitative results (or are at par) for most categories on all datasets while jointly learning dense correspondences. The qualitative comparisons with baselines 
highlight the efficiency of our approach. 

\vspace{-4mm}
\paragraph{Evaluation Metrics:}
\label{evaluation_metrics}
Correctly and efficiently evaluating the reconstruction quality has been a point of debate \cite{Tatarchenko2019WhatDS, achlioptas2017latent_pc, liu2019morphing, point_set_distance}. In this work, we evaluate the reconstruction quality by comparing the reconstructed shape with GT using (a) Chamfer distance (b) Earth Mover's Distance (EMD) and (c) Precision, Recall, F-score at 0.1 threshold. 


\section{Experimental Results}

\subsection{Qualitative and Quantitative Comparisons}





\paragraph{3D reconstruction on CUBS-200-2011:} We compare against CMR\cite{cmrKanazawa18} and SDF-SRN \cite{SDF-SRN} on the CUBS dataset in Figure~\ref{fig:cubs_comparison}. While SDF-SRN independently reconstructs each 3D object given the input image, CMR reconstructs each instance shape by deforming the category-specific mean mesh. On the other hand, TARS learns to reconstruct 3D instances implicitly by deforming the object space points to the canonical space. Compared to both deformation based reconstruction approaches, SDF-SRN generates noisy shapes \emph{(see noisy wings of the bird in row 2, Figure~\ref{fig:cubs_comparison})}. Credited to the implicit nature of our approach, the reconstructed shapes better respect the articulations of the GT objects (see rotated heads in Figure~\ref{fig:teaser}, open wings in Figure~\ref{fig:cubs_comparison} row 2).

\vspace{-5mm}
\paragraph{3D reconstruction on Shapenet:}
Figure~\ref{fig:shapenet_comparison} and Table~\ref{tab:shapenet-comparison} showcase qualitative and quantitative  comparison of the reconstructed shapes on the Shapenet dataset. As a mesh-based reconstruction algorithm, SoftRas \cite{liu2019softras} is able to reconstruct cars and planes, but fails on the chairs category, reason being the large intra-category topological variations. It fails to capture the details and only recovers the global shape underlying the input image. 
\sduggal{Being a neural implicit reconstruction approach, SDF-SRN \cite{SDF-SRN} captures both the global structure and the fine details. TARS 
matches the reconstructed shape fidelity of the SDF-SRN reconstructions, both quntitatively and qualitatively
\emph{(see the tail of the airplane, arms of the revolving chair in Figure~\ref{fig:shapenet_comparison})}, and also learns cross-instance structural correspondences for free. Thanks to the proposed higher-dimensional deformation field, our reconstructions respect the topology of the GT shape \emph{(both the arms of the couch in Figure~\ref{fig:shapenet_comparison} have holes in them)}.}

\begin{table}[t]
\centering
\scalebox{0.55}{
\begin{tabular}{llccccccc}
\specialrule{.2em}{.1em}{.1em}
Cat. & Method & \multicolumn{3}{c}{Chamfer $\downarrow$} & EMD $\downarrow$ & Precision $\uparrow$ & Recall & F score $\uparrow$\\
& & acc. & cov. & overall & & (\%) &  (\%) &  (\%) \\
\specialrule{.1em}{.05em}{.05em}
\rowcolor{blue!7} 
& SoftRas \cite{liu2019softras} & 0.372 & 0.302 & 0.337 & 0.723 & 93.04 & 96.62 & 94.80 \\
\rowcolor{blue!7} 
& SDF-SRN \cite{SDF-SRN} & \textbf{0.141} & 0.144 & 0.142 & 0.452 & \textbf{99.76} & 99.84 & \textbf{99.80}\\
\rowcolor{blue!7} 
\multirow{-3}{*}{Car} & TARS (ours) & \textbf{0.141} & \textbf{0.140} & \textbf{0.140}  & \textbf{0.446} & 99.70 & \textbf{99.81} & 99.75\\
\hline

\rowcolor{red!5} 
& SoftRas \cite{liu2019softras} & 0.572 & 0.475 & 0.523 & 1.017 & 82.56 & 89.18 & 85.74\\
\rowcolor{red!5}
& SDF-SRN \cite{SDF-SRN} & \textbf{0.352} & 0.315 & 0.333 & 0.854 & \textbf{94.18} & 95.21 & \textbf{94.69}\\
\rowcolor{red!5}
\multirow{-3}{*}{Chair} & TARS (ours) & \textbf{0.353} & \textbf{0.312} & \textbf{0.332} & \textbf{0.817} & 93.43 & \textbf{95.39} & 94.40\\
\rowcolor{green!2}
\hline

& SoftRas \cite{liu2019softras} & 0.215 & 0.207 & 0.211 & 0.588 & 98.74 & 98.42 & 98.58\\
\rowcolor{green!2}
& SDF-SRN \cite{SDF-SRN} & \textbf{0.193} & 0.154 & 0.173 & 0.576 & 98.55 & 99.11 & 98.83\\
\rowcolor{green!2}
\multirow{-3}{*}{Airplane} & TARS (ours) & \textbf{0.194} & \textbf{0.152} & \textbf{0.173} & \textbf{0.533} & \textbf{98.79} & \textbf{99.34} & \textbf{99.06}\\
\hline
\end{tabular}
}
\vspace{-2.5mm}
\caption{3D reconstruction results on ShapeNet. Compared to mesh based SoftRas algorithm, both the implicit approaches: SDF-SRN and our approach perform significantly better on all metrics.}
\label{tab:shapenet-comparison}
\vspace{-3.5mm}
\end{table}

\vspace{-4mm}
\paragraph{3D reconstruction on Pascal3D+:} The default Pascal3D+ dataset provides 2D-3D paired data by associating PASCAL VOC \cite{PascalVOC} and Imagenet \cite{Imagenet} images with the closest matching CAD model. Since, the same set of CAD models are used for both training and test set objects, generating object silhouettes (used both during training and inference) by rendering the 3D CAD models creates a bias between the train and the test sets. Thus, generalization results of prior reconstruction methods \cite{SDF-SRN, choy20163d} on the Pascal3D+ dataset should be taken with a grain of salt.  Unlike prior works \cite{SDF-SRN}, we demonstrate qualitative comparison on both the default biased Pascal3D+ dataset and an unbiased version of the same dataset. The main purpose to showcase results on both the default and the unbiased datasets is to dis-entangle the inherent limitations of prior works from the lack of generalization. Please refer to supp. dataset section for more details. 
We compare against CMR \cite{cmrKanazawa18} and SDF-SRN \cite{SDF-SRN} on three categories of Pascal3D+ dataset (cars, planes, and chairs).  As shown in Figure~\ref{fig:pascal_comparison_chairs}, CMR \cite{cmrKanazawa18} suffers significantly on the chairs category and even fails to capture the global shape, the reason being their mesh representation which doesn't allow breaking the initial mesh connectivity (/ topology). Even on the planes category of both the default dataset (supp. Figure~\ref{fig:supp_pascal_comparison_airplanes_default}) and the unbiased dataset (Figure~\ref{fig:pascal_comparison_planes}, supp. Figure~\ref{fig:supp_pascal_comparison_planes_unbiased}, supp. Figure~\ref{fig:supp_pascal_comparison_planes_unbiased_v1}), CMR fails to capture the details and rather generates self-intersecting and similar-looking meshes for different plane instances. 
SDF-SRN \cite{SDF-SRN} does capture the overall shape details well. However, because of the challenging nature of the real-world images, compared to its performance on Shapenet dataset \cite{Chang2015ShapeNetAI}, it under-performs and generate much noisier reconstructions on the real-world Pascal dataset \emph{(see noise on the reconstructed planes in Figure~\ref{fig:pascal_comparison_planes}, ripples on the reconstructed SDF-SRN cars in supp. Figure ~\ref{fig:supp_pascal_comparison_cars_default},  \emph{noisy} reconstructed sofa in Figure~\ref{fig:pascal_comparison_chairs})}. Further, it fails to maintain the topological details of the GT 3D shapes \emph{(eg: lack of details on the back of the chair in row 3, missing arms of chairs in row 2, 3 of Figure~\ref{fig:pascal_comparison_chairs})}. In comparison, as can be seen from the results on both the default dataset (Figure~\ref{fig:pascal_comparison_chairs}, supp. Figure~\ref{fig:supp_pascal_comparison_cars_default}, ~\ref{fig:supp_pascal_comparison_airplanes_default}, ~\ref{fig:supp_pascal_comparison_chairs_default}) and the unbiased dataset (supp. Figure~\ref{fig:supp_pascal_comparison_planes_unbiased}, ~\ref{fig:supp_pascal_comparison_planes_unbiased_v1}), our reconstructions are (a) much less noisier, (b) respect the topology of the underlying shapes, and (c) better captures both the global shape and the finer details. 

\begin{figure}[t]
\centering
\includegraphics[width=0.72\linewidth]{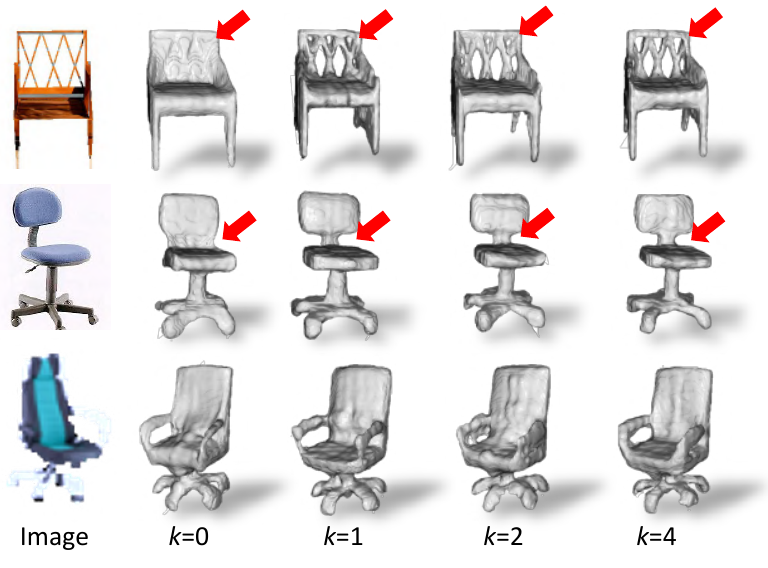}
\vspace{-3mm}
\caption{Ablation of dimensionality ($k$+3) of deformation field on Pascal3D+ (default) chairs. $k$ equals the dimensionality of the additional point features.}
\label{fig:ablation_deformation_field}
\vspace{-4mm}
\end{figure}

\vspace{-4mm}
\paragraph{3D reconstruction on Pix3D Chairs:} To showcase the generalization capability of our proposed approach, we demonstrate qualitative comparisons on the Pix3D chairs dataset. Figure~\ref{fig:pix3d_shapenet} compares shapenet-trained SDF-SRN\cite{SDF-SRN} and our approach on the Pix3D test split. Despite the challenging nature of Pix3D dataset (diverse 3D shapes, variable texture, material and environment conditions), both the approaches generalizes well from the synthetic Shapenet dataset to the real-world Pix3D dataset. Further, thanks to the topologically-aware deformation field, our approach maintains the topological structure of the GT shapes (see reconstructed holes in chairs in row 2, 3 of Figure~\ref{fig:pix3d_shapenet}). Please refer to supp. for comparison of reconstruction approaches trained only on real-world chairs (Pix3D train + Pascal3D). 

\vspace{-4mm}
\paragraph{Learned deformation field:} We visualize our learned deformation fields in Figure~\ref{fig:teaser}, ~\ref{fig:cubs_comparison} and supp. Figure~\ref{fig:supp_chairs_deformation_vis}. The color codes denote the corresponding canonical 3D points obtained by mapping the 3D object space points to the canonical space using the learned deformation fields. 
Our deformation field consistently learns to deform similar object parts to similar regions of the learned mean shape \emph{without any form of part supervision} (eg: legs of all chairs in supp. Figure~\ref{fig:supp_chairs_deformation_vis} are consistently painted similarly with yellow, green, blue and pinkish-white). \emph{Similar deformation consistency is observed in the cubs category, despite the structural and non-rigid articulation dependent variations (see Figure~\ref{fig:cubs_comparison}).} This validates that deformation-based approaches can inherently learn category-specific structural relations \emph{(without any supervision)} leveraging just single-view image collections.

\vspace{-4mm}
\paragraph{Leveraging deformation fields for texture transfer:} We showcase the utility of the learned deformation field for the task of texture transfer in Figure~\ref{fig:teaser}. We first manually paint a 3D mesh and then transfer the painted texture to other meshes using the learned deformation field of the two meshes. As can be seen in the figure, structurally similar parts of both the source and the target meshes are painted similarly. The checkered stripe patterns and the parallel stripe patterns of the source meshes of row 2 and 3 respectively, are maintained in the target meshes, highlighting the structural details captured by the learned deformation fields.


\subsection{Ablation Study}
We ablate the efficiency of the learned point features on Pascal3D+ categories in Table~\ref{tab:pascal-point-features-ablation} and Figure ~\ref{fig:ablation_deformation_field}. Despite the bias in Pascal3D+ default dataset, such an ablation is useful as it helps understand the inherent limitation of the implicit deformation approaches, by ruling out the lack of generalization as a potential factor for the lack of fidelity. As can be seen from the table, learning higher-dimensional deformation field leads to considerable improvement in chamfer coverage and EMD metric (while still improving chamfer accuracy metric). 
\emph{This highlights that point features are contributing in the enhancement of details and structures present in GT shapes, and are thus crucial for reconstructing topologically varying categories.} Figure~\ref{fig:ablation_deformation_field} qualitatively validates this fact. We didn’t observe significant improvements for
$k > 4$, where $k$ equals point features dimensionality. 

\begin{table}[t]
\centering
\scalebox{0.50}{
\begin{tabular}{llccccccc}
\specialrule{.2em}{.1em}{.1em}
Category & Method & \multicolumn{3}{c}{Chamfer $\downarrow$} & EMD $\downarrow$ & Precision $\uparrow$ & Recall & F score $\uparrow$\\
& & acc. & cov. & overall & & (\%) &  (\%) &  (\%) \\
\specialrule{.1em}{.05em}{.05em}
\rowcolor{blue!7} 
& Ours (w/o point features) & 0.379 & 0.473 & 0.426 & 0.949 & 96.97 & 91.77 & 94.30\\
\rowcolor{blue!7} 
\multirow{-2}{*}{Car} & Ours & \textbf{0.363} & \textbf{0.386} & \textbf{0.374} & \textbf{0.763} & \textbf{97.00} & \textbf{95.63} & \textbf{96.31}\\

\rowcolor{red!5}
& Ours (w/o point features) & 0.539 & 0.485 & 0.512 & 1.291 & 86.95 & 91.44 & 89.14\\
\rowcolor{red!5}
\multirow{-2}{*}{Chair} & Ours & \textbf{0.527} & \textbf{0.426} & \textbf{0.476} & \textbf{1.171} & \textbf{89.75} & \textbf{94.83} & \textbf{92.22}\\

\rowcolor{green!2}
& Ours (w/o point features) & 0.576 & 0.561 & 0.568 & 1.341 & 87.77 & 93.16 & 90.38\\
\rowcolor{green!2}
\multirow{-3}{*}{Airplane} & Ours & \textbf{0.547} & \textbf{0.530} & \textbf{0.538} & \textbf{1.302} & \textbf{88.50} & \textbf{93.76} & \textbf{91.05}\\
\hline
\end{tabular}
}
\vspace{-2.5mm}
\caption{Point features ablation on Pascal 3D+ chairs dataset.} 
\label{tab:pascal-point-features-ablation}
\vspace{-6mm}
\end{table}

%% file: tex_files/conclusion.tex
\section{Conclusion}

In this work, we presented an approach which can learn to reconstruct 3D shapes, given a (category-specific) collection of unpaired 2D images. The proposed approach, TARS, tackles the problem of single-view reconstruction by implicitly learning to deform different object instances to a learned category specific mean shape. By transforming the 3D deformation field to a higher-dimensional field, we corroborated that the learned-deformation field is topologically-aware. As a result, our reconstructed shapes capture the global structure of the underlying GT shape and also resembles the GT shapes much than than the prior works in terms of fine structural and topological details. Furthermore, the learned deformation field implicitly captures the structural properties of the category, without any explicit supervision. 
Overall, our results represent an encouraging step towards generalization of reconstruction systems to the internet of images.

%% file: tex_files/supp.tex

\section*{Appendix}

We first provide additional experimental details, which include dataset
details, implementation details and baselines’ descriptions. Next, we provide additional experimental results in terms of further qualitative and quantitative analysis. Finally, we discuss the potential future works.

\section{Additional Experimental Details}

\subsection{Dataset Details} We performed experiments on the following datasets: synthetic Shapenet \cite{Chang2015ShapeNetAI}, real-world Pascal 3D+ \cite{Pascal3D}, real-world CUB-200-2011 \cite{WelinderEtal2010} and real-world Pix3D chairs \cite{pix3d}:

\vspace{-2mm}
\paragraph{Shapenet:} We used car, planes and chairs category of the Shapenet v2 dataset \cite{Chang2015ShapeNetAI} for our experiments. Shapenet Image dataset is generated by rendering the synthetic CAD models from different sampled viewpoints. Shapenet CAD models are associated with texture files (generally possessing only diffuse texture component). For all experiments, we followed SDF-SRN's data settings (mentioned in their paper section 4.1 and Appendix section B.1). To summarize: we used \emph{2830 training, 809 validation and 405 test CAD model for airplanes, 2465 training, 359 validation and 690 test CAD models for cars category and 4744 training, 678 validation and 1356 test CAD objects for chairs category.} 

\vspace{-2mm}
\paragraph{Pascal3D+:} Pascal3D+ \cite{Pascal3D} is a dataset of real-world camera images with annotated 3D CAD models. Compared to Shapenet, Pascal3D+ data is challenging as camera images are captured in real-world scenarios with variable lightning conditions, variable object occlusions, diverse object textures etc. We follow SDF-SRN's data settings for Pascal3D+ dataset (mentioned in their paper \cite{SDF-SRN} section 4.2 and appendix section B.2). To summarize the data-splits: \emph{we used 991 training and 974 validation examples for airplane category, 2847 training and 2777 validation examples for cars category, and 539 training and 514 validation examples for chairs category.} The object silhouettes used both during training and test phases to mask out the foreground object RGB from the background are generated by rendering a fixed set of CAD models. \emph{Since the same CAD models are used to generate object silhouettes both during training and testing, the dataset posseses a bias (highlighted in Tulsiani \etal \cite{TulsianiZEM17} Appendix A2.2).} While prior work \cite{choy20163d, SDF-SRN} showcase generalization results only on this biased dataset, we address this issue by using silhouette masks generated by an off-the-shelf instance segmentation network \cite{LiHM15} as done by Tulsiani \etal \cite{TulsianiZEM17}. 
Results on the unbiased Pascal3D+ planes dataset are shown in Figure~\ref{fig:supp_pascal_comparison_planes_unbiased}. For results on unbiased chairs dataset, refer to Pix3D results (Figure~\ref{fig:supp_pix3d_shapenet_comparison}, ~\ref{fig:supp_pix3d_comparison}).

\vspace{-3mm}
\paragraph{Pix3D chairs dataset:} Similar to Pascal3D+ \cite{Pascal3D} dataset, Pix3D dataset \cite{pix3d} is a real-world dataset, containing 2D image to 3D CAD model mappings. However, unlike Pascal3D+ dataset, (a) the 3D CAD models align better with the 2D images, (b) different set of CAD models are used for training and test set images, therefore the dataset is unbiased. Some images of the Pix3D chairs dataset are highly occluded/ truncated. We  removed such images using the annotated truncation tag associated with each image and also by manual filtering. Overall, we used 2196 Pix3D chair images for training and 637 chair images for test. Since the overall dataset is significantly small (compared to tens of thousands of images in shapenet dataset), we augment Pix3D training set with the 539 Pascal3D training chair images. To highlight, \emph{this dataset is still significantly smaller} considering the amount of variations (in terms of light, material, texture) present in the real-world image collection. Also, since each CAD model is rendered at multiple viewpoints to generate a 2D-3D mapping, the overall 3D information used to train the reconstruction networks is much smaller.

\vspace{-3mm}
\paragraph{CUB-200-2011 dataset:} We used the annotated CUBS dataset released alongside the CMR \cite{cmrKanazawa18} codebase. Overall, the training set has 5964 images and the test set has 2874 images. Each image is associated with a silhouette map and a weak-perspective camera pose generated using 2D annotated keypoints and SFM registration of the keypoints. Please refer to CMR \cite{cmrKanazawa18} (section 3.1) for more details.

\subsection{Implementation Details} Our deformable reconstruction pipeline (as shown in paper Figure 2) consists of \emph{DeformNet}, \emph{Canonical Shape Generator}
and (an LSTM-based) \emph{Differentiable Renderer} modules. We need to ensure that each module performs its desired task, despite the lack of explicit supervision. Simply jointly training the three modules fails to ensure that, and hence results in poor performance. In order to effectively train these modules, we follow a curriculum learning strategy. We split the training phase into two main stages and two intermediate pre-training stages: \\

\vspace{-2mm}
\textbf{In Stage 1}, we directly learn to reconstruct the 3D shape (in form of signed-distance field) given the corresponding input image captured from a known input viewpoint. We adopt SDF-SRN \cite{SDF-SRN} for this task and \emph{train the shape generator module \footnote{In Stage 1, the shape generator module is trained to reconstruct any 3D shape given the corresponding camera image. In stage 2, this shape generator module is fine-tuned for reconstructing only the canonical shape and is thus termed as Canonical Shape Generator.}, image encoder and the differentiable renderer module in this stage}. Given an input image, we first map it to a latent-code using Imagenet pre-trained Resnet encoder \cite{He2015}. The latent-code is used by a hyper-network to generate the weights of the shape generator module. The shape generator module then learns to map any 3D point in the object space to its corresponding SDF value. The differentiable renderer module 
is also trained alongside to render the learned geometry from the given input viewpoint. It is an LSTM module which takes as input the intermediate-level features from the shape generator module (corresponding to the sampled 3D point) and predicts the ray marching step along the input ray direction. Using the 3D point and the ray-marching step, the next point along the input ray direction is generated. The above mentioned procedure is then repeated for a fixed number of ray-marching steps.
In order to ensure that shape generator module can operate on higher-dimensional input (3D point + point features) in Stage 2, we additionally pass ``un-conditioned'' point-features as input to the shape generator module. These point-features (4-dimensional in our experiments) are generated by simply passing the input 3D points through a two-layer MLP (which is not conditioned on the input-image). Instead of passing the 3D points plus the point features to the shape generator network, we pass the concatenation of 3D points, their position encoding and the positional encoding of the point features as input to the network. \\
    
\textbf{In Stage 2}, \emph{we train the DeformNet module, while fine-tuning the image encoder, shapenet generator module and the differentiable render.} The DeformNet module takes as input a 3D point in the object space and maps it to a higher-dimensional (7-dimensional in our experiments) canonical point (3D point deformation + 4D object-space point features) using the learned higher dimensional deformation field. Alongside predicting the deformation field, it also predicts the view-independent RGB value for the input 3D point. Next, given the higher-dimensional canonical point, the shape generator module learns to predict the corresponding SDF value. In stage 2, the shape generator module only focuses on reconstructing the canonical 3D shape, and hence is termed as the canonical shape generator. The differentiable renderer in stage 2 takes as input the object-space features of the sampled 3D point (sampled along the input ray) as learned by the DeformNet.  Like stage 1, weights of both DeformNet and Canonical Shape Generator are learned through hyper-networks. Unlike Stage 1, where the hyper-network for the shape generator is conditioned on the input image latent-code, the weights of the canonical shape generator are predicted by a hyper-network conditioned on a canonical shape latent-code (which is optimized jointly). The canonical shape-latent code is initialized by the mean of all training images' latent codes predicted in Stage 1. Like the shape generator, the input to the Deformnet is the concatenation of the 3D point and its positional encoding. \\ 


\vspace{-2mm}
\textbf{Pre-training phases:} Following SDF-SRN \cite{SDF-SRN}, prior to Stage 1 training, we first pre-train the shape reconstruction module to predict the SDF-space of a zero centered 3D sphere (conditioned on random latent code in place of image-based latent code used in Stage 1). This helps the network better learn the 3D object signed-distance fields in stage 1. Prior to Stage 2 (and post stage 1 training), we overfit the DeformNet module to deform points belonging to the initial canonical space (SDF space generated using initial canonical shape-latent code and the pre-trained shape generator module) to a 3D sphere, such that SDF of the initial point in the canonical space is equal to the SDF of the deformed point w.r.t the 3D sphere. \\ 

\vspace{-2mm}
\textbf{Architecture details:} We now provide the architecture details for the three modules: The shape generator module is implemented as an MLP with two output heads, one used to predict the SDF value for the input 3D point and the other used to predict the point's RGB value (during stage 1). The shared MLP backbone between the two output heads has multiple linear layers with LayerNorm and ReLU activation,
while the output heads are just linear layers. The weights and the biases for each layer are generator by different hyper-networks, which themselves are MLPs. The high-level architecture is adopted from SDF-SRN \cite{SDF-SRN}. The architecture for the Deformnet is similar to the shape generator module, with three output heads learning 3D point deformation, 4D point features and the RGB value for each input 3D point.
For the LSTM module of the differentiable renderer \cite{SRN}, we kept the output and the hidden state dimension to be 32. \\




\vspace{-2mm}
\subsection{Baselines}

\paragraph{SDF-SRN \cite{SDF-SRN}:} We directly used the open-sourced codebase and pretrained models of SDF-SRN \cite{SDF-SRN} to generate the results. Note, the open-sourced pre-trained Pascal3D+ models belong to the default biased Pascal3D+ dataset.

\vspace{-2mm}
\paragraph{SoftRas \cite{liu2019softras}:} We used the released codebase and trained the SoftRas shape reconstruction model on the Shapenet dataset. For fair comparison, we commented out the multi-view consistency loss in the open-source implementation and rather rendered the reconstructed mesh only at the source viewpoint to supervise the training pipeline.

\vspace{-2mm}
\paragraph{CMR \cite{cmrKanazawa18}:} We used the released codebase for training CMR \cite{cmrKanazawa18} on Pascal3D+ dataset. For fair comparison, we did not use the key-point loss and only used the RGB and the silhouette loss for supervising the pipeline. Prior to training, we initialized their mean shape prior using the 3D mesh template shared alongside the CMR codebase. 
For chairs category, because of the large intra-category topological variations, we found out that using a template mesh (which was not isomorphic to sphere) for initialization of the mean shape leads to poor training and hence poor reconstructions. Therefore, (following SDF-SRN \cite{SDF-SRN}) we used a 3D sphere to initialize the mean shape for Pascal3D+ chairs reconstruction task.

\begin{figure}[t]
\centering
\includegraphics[width=\linewidth]{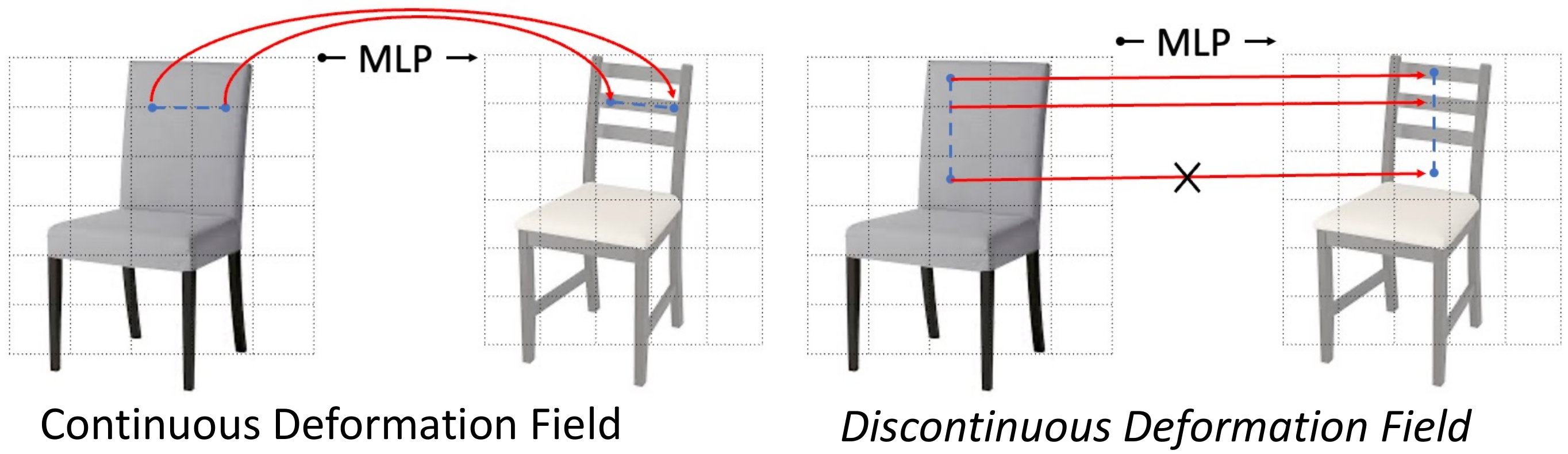}
\vspace{-5mm}
\caption{\textbf{Discontinuities in a deformation field:} (Left) Example of continuous mapping of a set of 3D points from source chair to target chair. (Right) Example of dis-continuous mapping from source to target chair.}
\label{fig:supp_topological_deformation_example}
\vspace{-5mm}
\end{figure}

\subsection{Metrics} We used symmetric Chamfer distance (CD) and Earth Mover's distance (EMD) to quantitatively measure the fidelity of the reconstructed meshes. CD is defined as the sum of squared distance of each 3D point on the ground-truth shape ($\mathbf{X}$) to the closest surface point on the reconstructed shape ($\mathbf{Y}$) and vice-versa.

\begin{align}
    \mathrm{CD}(\mathbf{X}, \mathbf{Y}) =  \frac{1}{2|\mathbf{X}|}\sum_{x \in \mathbf{X}} \min_{y \in \mathbf{Y}} \| x-y \|_{2} \vspace{-2mm} \;+\; \notag \\  \frac{1}{2|\mathbf{Y}|}\sum_{y \in \mathbf{Y}} \min_{x \in \mathbf{X}} \| x-y \|_{2} \vspace{-2mm} \notag
\end{align}

EMD is defined as the sum of the squared distance of each point in the GT point cloud ($\mathbf{X}$) to its bijective mapping in the reconstructed point cloud.

\[\mathrm{EMD}(\mathbf{X}, \mathbf{Y}) = \min_{\phi: X \rightarrow Y} \sum_{x \in X} \|x - \phi(x)\|_2 \]

We also used Precision and Recall as robust alternatives \cite{Tatarchenko2019WhatDS} to chamfer distance.

\vspace{-2mm}
\[\mathrm{Precision(\mathbf{X}, \mathbf{Y})} = \frac{1}{|\mathbf{Y}|}\sum_{y \in \mathbf{Y}} \bigg[ \min_{x \in\mathbf{X}} \|x-y\|_{2} <= t \bigg]\]

\[\mathrm{Recall(\mathbf{X}, \mathbf{Y})} = \frac{1}{|\mathbf{X}|}\sum_{x \in \mathbf{X}} \bigg[ \min_{y \in\mathbf{Y}} \|x-y\|_{2} <= t \bigg]\] \\

We set the true-positive threshold to 0.1.

\begin{figure}[t]
\centering
\includegraphics[width=0.90\linewidth]{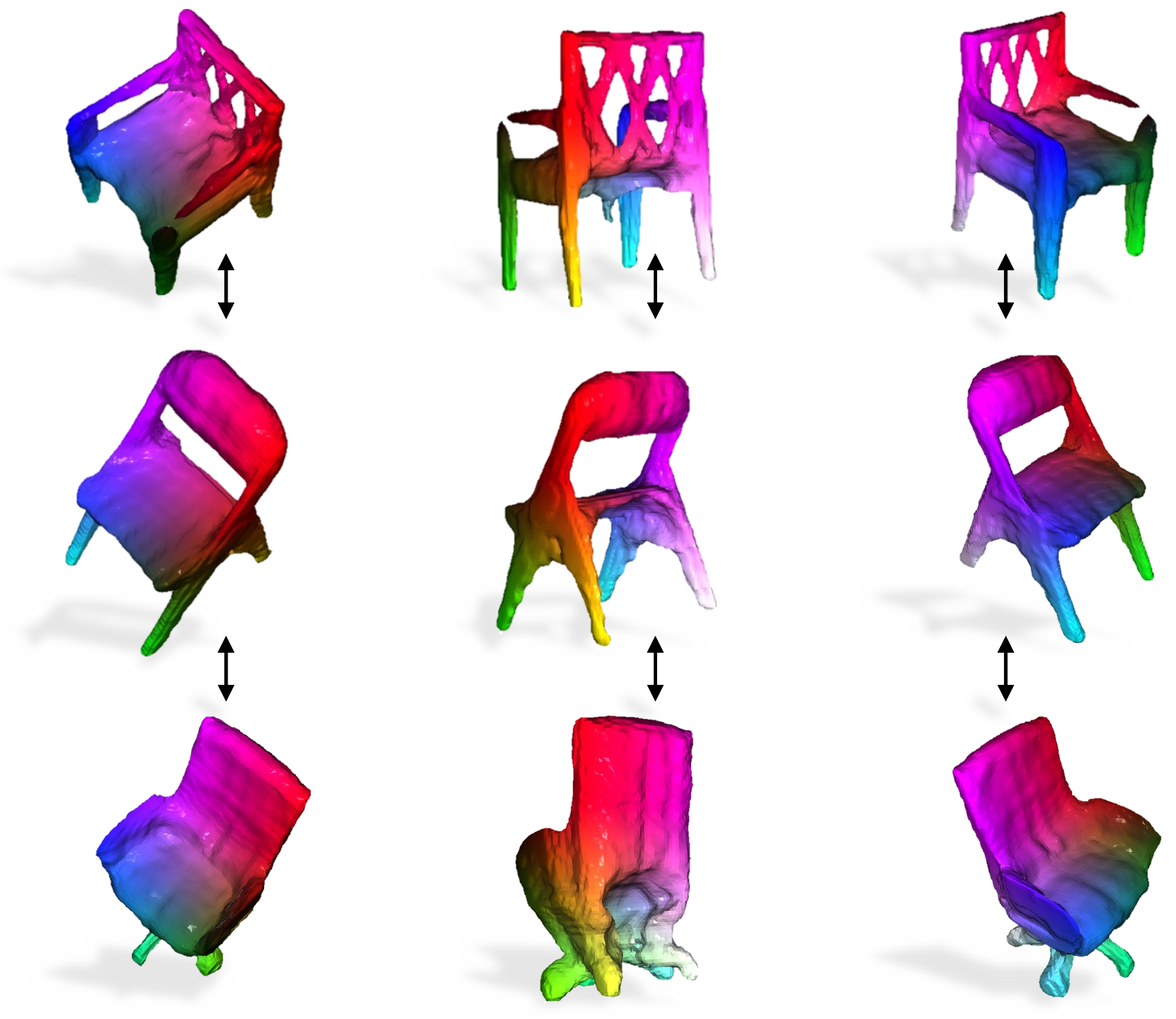}
\vspace{-2mm}
\caption{\textbf{Visualization - Topologically-aware deformation field:} The learned deformation fields map similar structures/ parts of different object instances to similar canonical space regions.}
\label{fig:supp_chairs_deformation_vis}
\vspace{-2mm}
\end{figure}

\section{Analysis: Topologically-Aware Deformation Field} \label{sec:topological_understanding}
Figure ~\ref{fig:supp_topological_deformation_example} motivates the need of the topologically-aware deformation field. Our approach reconstructs the target shape by mapping 3D points from the target space to the source space using the learned deformation field. Since the deformation field is learned implicitly using an MLP, the inductive continuous nature of the MLP would deform 3D points continuously from the target space to the source space. Thus such a deformation field can truly reconstruct the target shape from the source shape only when both the shapes are of similar topologies (Figure ~\ref{fig:supp_topological_deformation_example} left). To over come this topological restriction, we take inspiration from Level Set Methods and learn additional per-point features which potentially guide the network on how to modify the deformation field to truly reconstruct the target shape from the source.

\section{Additional Experimental Results}

\subsection{Qualitative Analysis}

\paragraph{Single-View 3D Reconstruction on Shapenet:}
Figure~\ref{fig:supp_shapenet_comparison_cars}, Figure~\ref{fig:supp_shapenet_comparison_chairs} and Figure~\ref{fig:supp_shapenet_comparison_planes} compare our proposed approach with SDF-SRN \cite{SDF-SRN} baseline on the Shapenet dataset's cars, planes and chairs category respectively. Compared to the mesh-based baseline (SoftRasterizer \cite{liu2019softras}, as shown in main-paper Figure 6) both our proposed approach and SDF-SRN \cite{SDF-SRN} perform significantly well, thanks to the use to neural implicit modeling. 
Thanks to the inherently learned category-specific structural priors (inherent property of deformable models) our shapes have fewer artifacts compared to SDF-SRN's shapes \emph{(see SDF-SRN's noisy reconstructions in row 1 chair 1, row 2 chair 2(sofa), row 5 chair 1 of Figure~\ref{fig:supp_shapenet_comparison_chairs})}.

\begin{figure*}
\centering
\includegraphics[width=\linewidth]{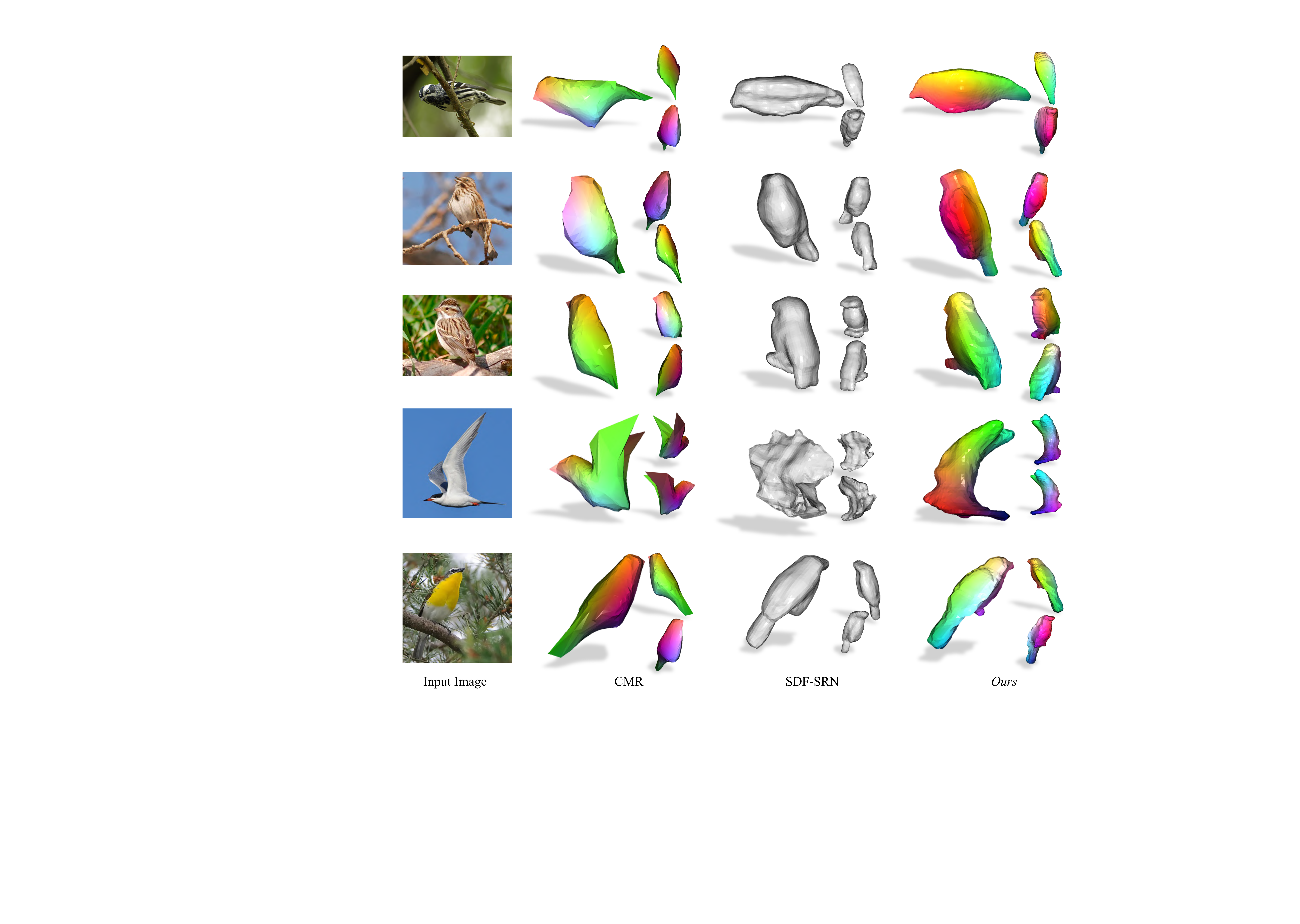}
\vspace{-2mm}
\caption{\textbf{3D Reconstruction on CUB-200-2011 from Single 2D Image}}
\label{fig:supp_cub_comparison}
\vspace{-5mm}
\end{figure*}

\begin{figure*}
\centering
\includegraphics[width=\linewidth]{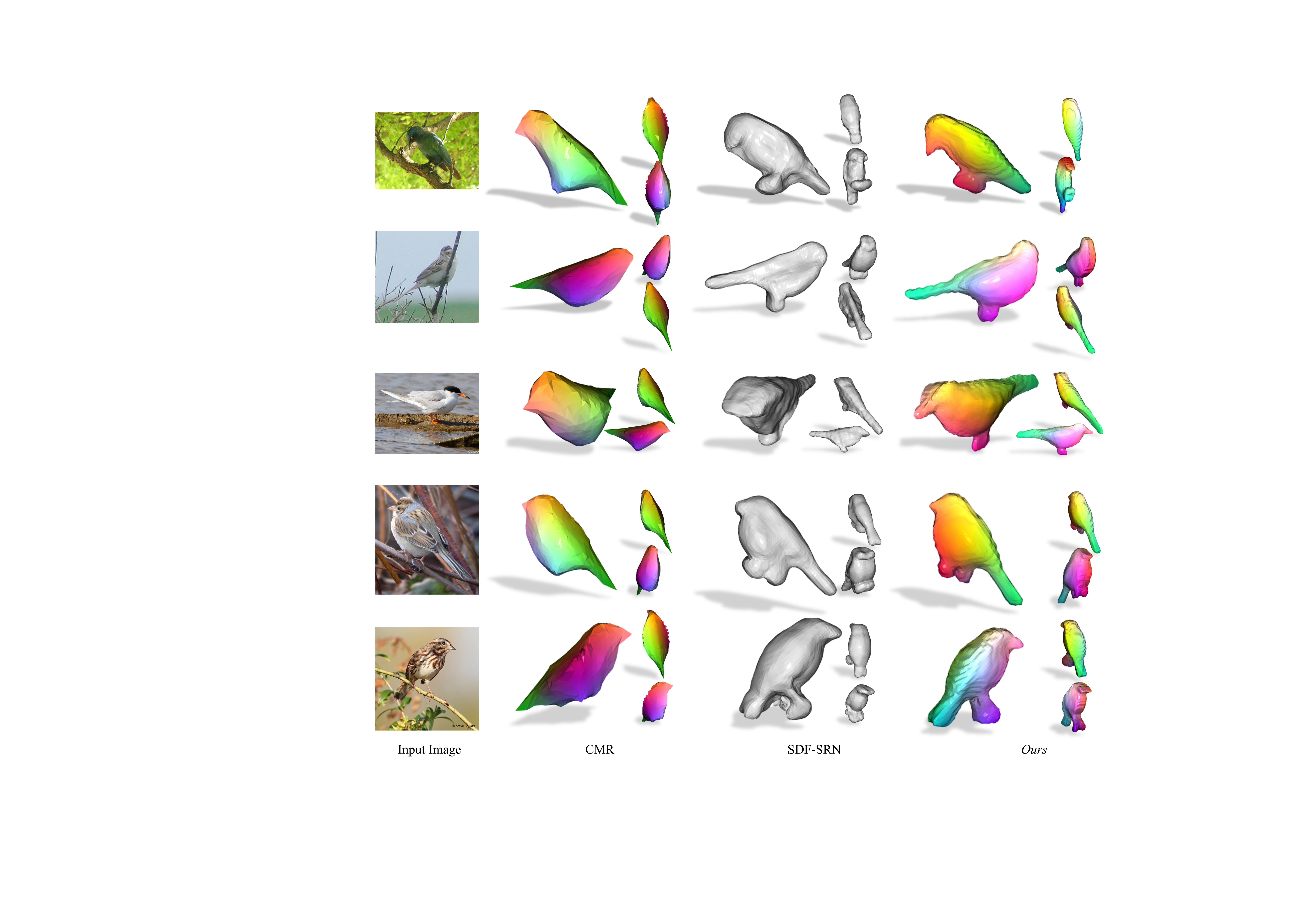}
\vspace{-2mm}
\caption{\textbf{3D Reconstruction on CUB-200-2011 from Single 2D Image}}
\label{fig:supp_cub_comparison_v1}
\vspace{-5mm}
\end{figure*}

\vspace{-3mm}
\paragraph{Single-View 3D Reconstruction on CUBS-200-2011:} Figure~\ref{fig:supp_cub_comparison} and Figure~\ref{fig:supp_cub_comparison_v1} showcase additional qualitative comparisons between SDF-SRN\cite{SDF-SRN} and our approach on the CUBS-200-2011 dataset. Our approach showcase consistent improvement over prior works in terms of (a) less-noisy reconstructions (row 3 right, row 4 left), (b) better articulations (row 2 left, row 4 right) and (c) better capture of overall geometric structure (reconstructed beaks, foots, legs etc).

\vspace{-3mm}
\paragraph{Single-View 3D Reconstruction on Pascal3D+ (default):}
Figure~\ref{fig:supp_pascal_comparison_cars_default}, Figure~\ref{fig:supp_pascal_comparison_airplanes_default} and Figure~\ref{fig:supp_pascal_comparison_chairs_default} compare our proposed approach with SDF-SRN \cite{SDF-SRN} baseline on the Pascal3D+ dataset's cars, planes and chairs category respectively. Compared to Shapenet, Pascal3D+ is a challenging real-world dataset with object images having diverse textures, captured under variable environmental (lightning) conditions and under variable nature of object occlusion. As a result, SDF-SRN reconstructions on Pascal3D+ are much noisier compared to their results on Shapenet \emph{(see ripples on car surfaces in Figure~\ref{fig:supp_pascal_comparison_cars_default} and noisy reconstructed chairs in Figure ~\ref{fig:supp_pascal_comparison_chairs_default})}. By learning to deform all object instances to a particular category-level canonical shape, we are able to regularize the reconstruction procedure and hence generate smoother shapes with 
much fewer artifacts. Moreover, compared to SDF-SRN \cite{SDF-SRN}, our reconstructions are able to capture the finer shape details captured in the input image \emph{(eg: reconstructed front wheel on planes in row 1, row 2 and row 4, reconstructed plane propeller in row 6 of Figure ~\ref{fig:supp_pascal_comparison_airplanes_default})}. Our shapes also maintain the topological details of the GT shape underlying the input image \emph{(see chairs in Figure ~\ref{fig:supp_pascal_comparison_chairs_default})}.

\begin{figure*}
\centering
\includegraphics[width=0.9\linewidth]{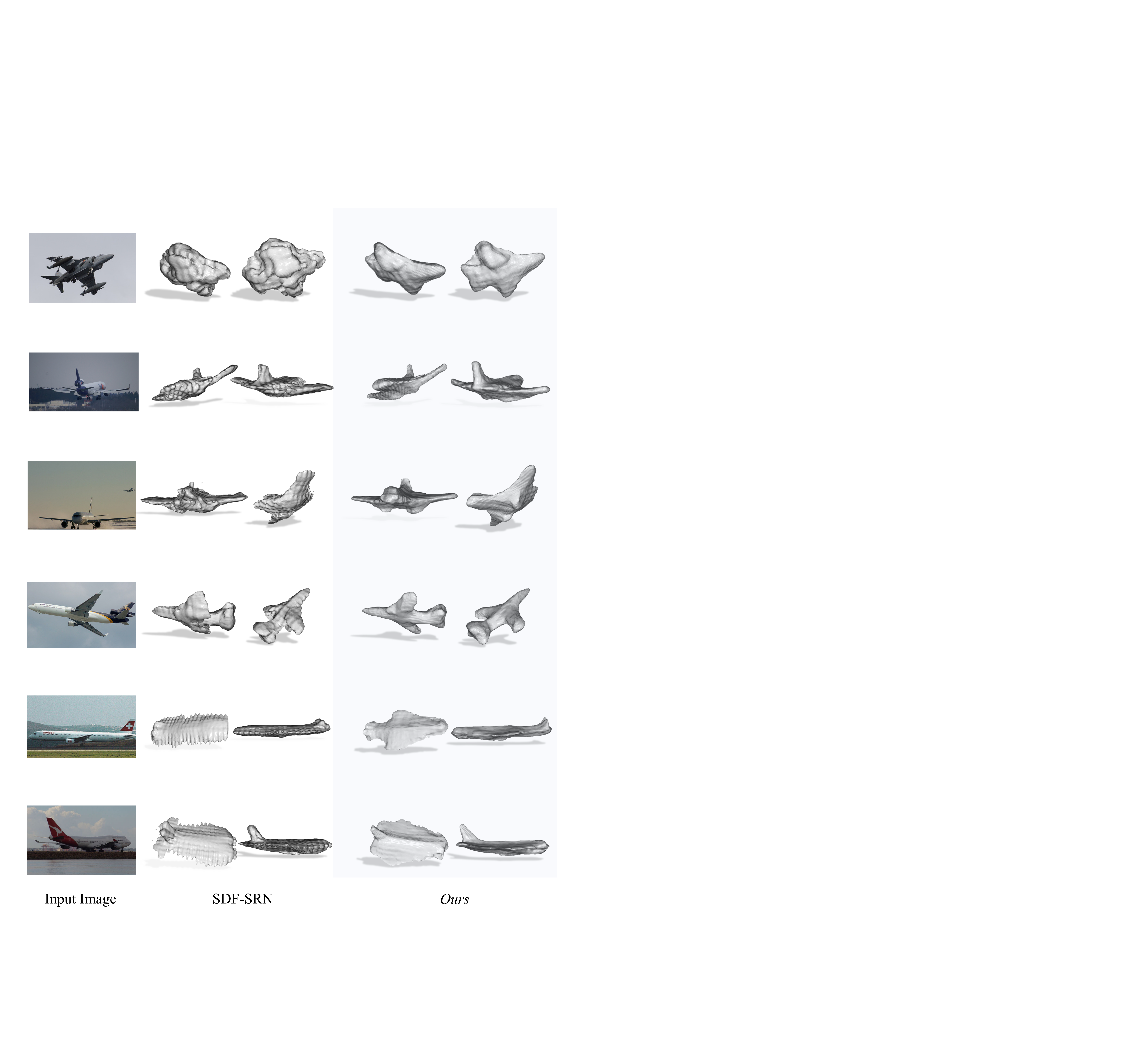}   
\vspace{-2mm}
\caption{\textbf{3D Reconstruction on Pascal3D+ (unbiased) Airplanes from Single 2D Image}}
\label{fig:supp_pascal_comparison_planes_unbiased}
\vspace{-5mm}
\end{figure*}

\begin{figure*}
\centering
\includegraphics[width=0.9\linewidth]{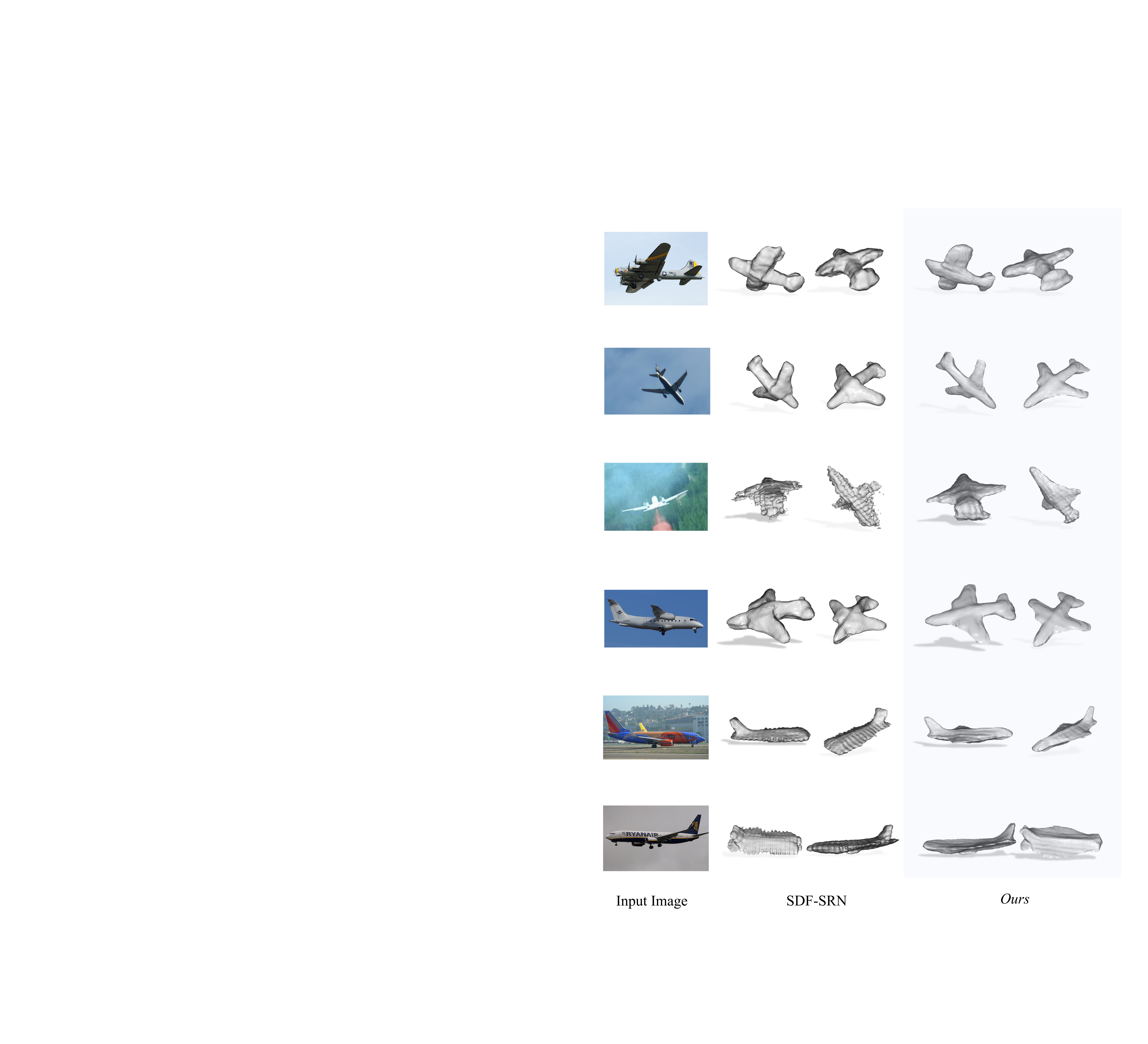}   
\vspace{-2mm}
\caption{\textbf{3D Reconstruction on Pascal3D+ (unbiased) Airplanes from Single 2D Image}}
\label{fig:supp_pascal_comparison_planes_unbiased_v1}
\vspace{-5mm}
\end{figure*}

\begin{figure*}
\centering
\includegraphics[width=0.90\linewidth]{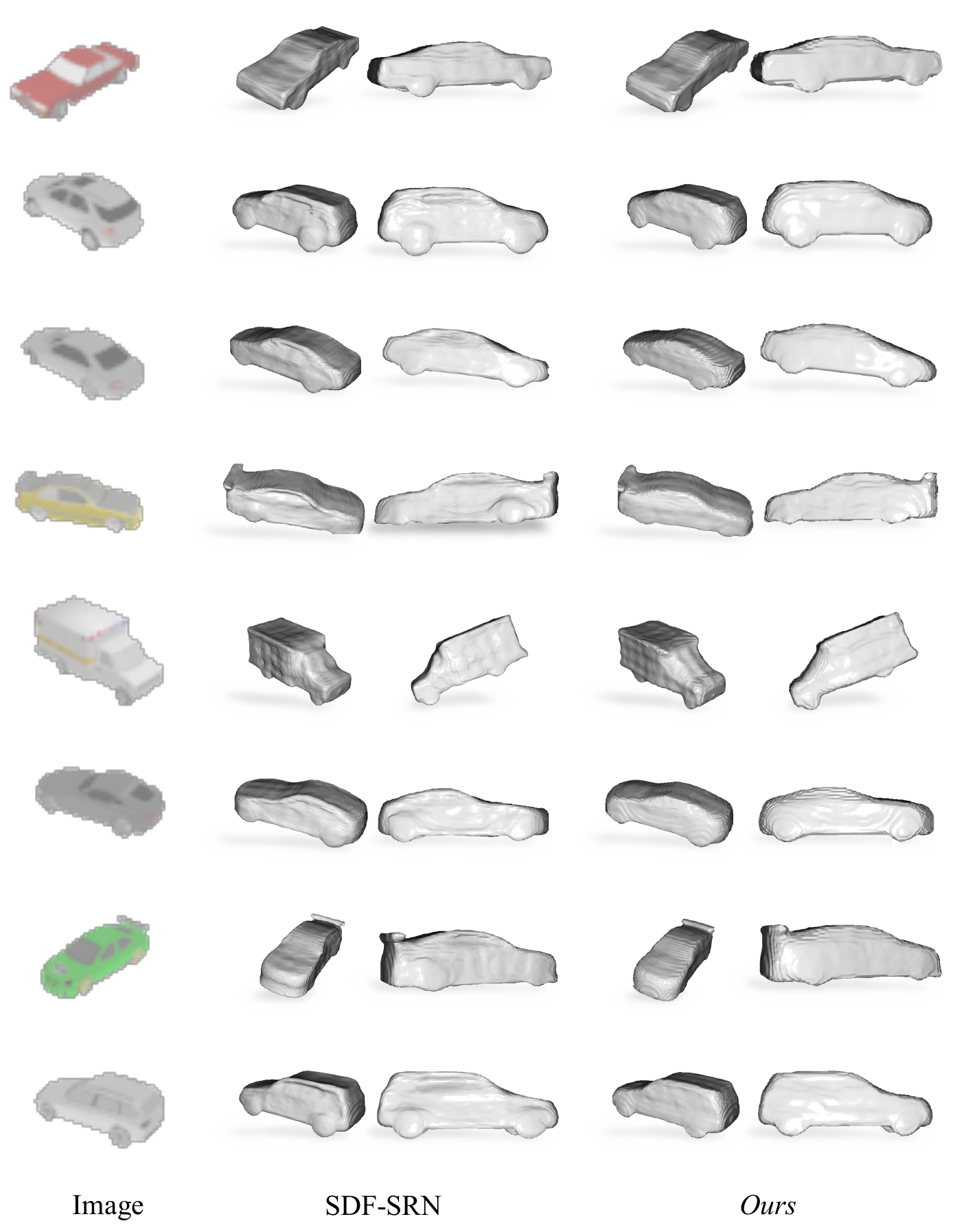}
\vspace{-2mm}
\caption{\textbf{3D Reconstruction on Shapenet Cars from Single 2D Image} Our approach amtches the shape fidelity of SDF-SRN while leveraging cross-instance correspondences for free.}
\label{fig:supp_shapenet_comparison_cars}
\end{figure*}

\begin{figure*}
\vspace{-8mm}
\centering
\includegraphics[width=0.70\linewidth]{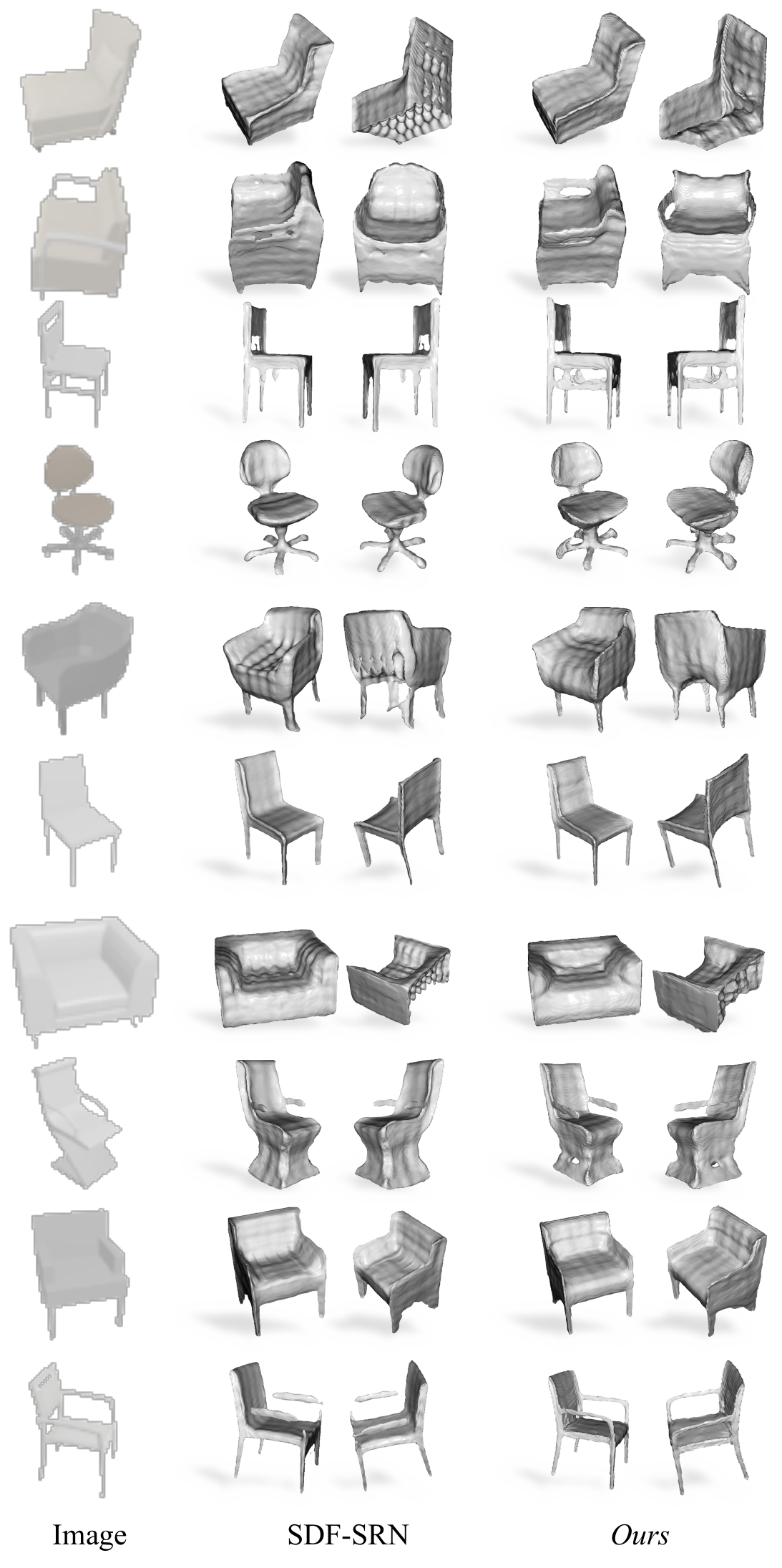}
\vspace{-2mm}
\caption{\textbf{3D Reconstruction on Shapenet Chairs from Single 2D Image} As can be seen, our reconstructions are less noiser, thanks to the learned deformation field which acts as a regularizer.}
\label{fig:supp_shapenet_comparison_chairs}
\vspace{-5mm}
\end{figure*}

\begin{figure*}
\centering
\includegraphics[width=0.9\linewidth]{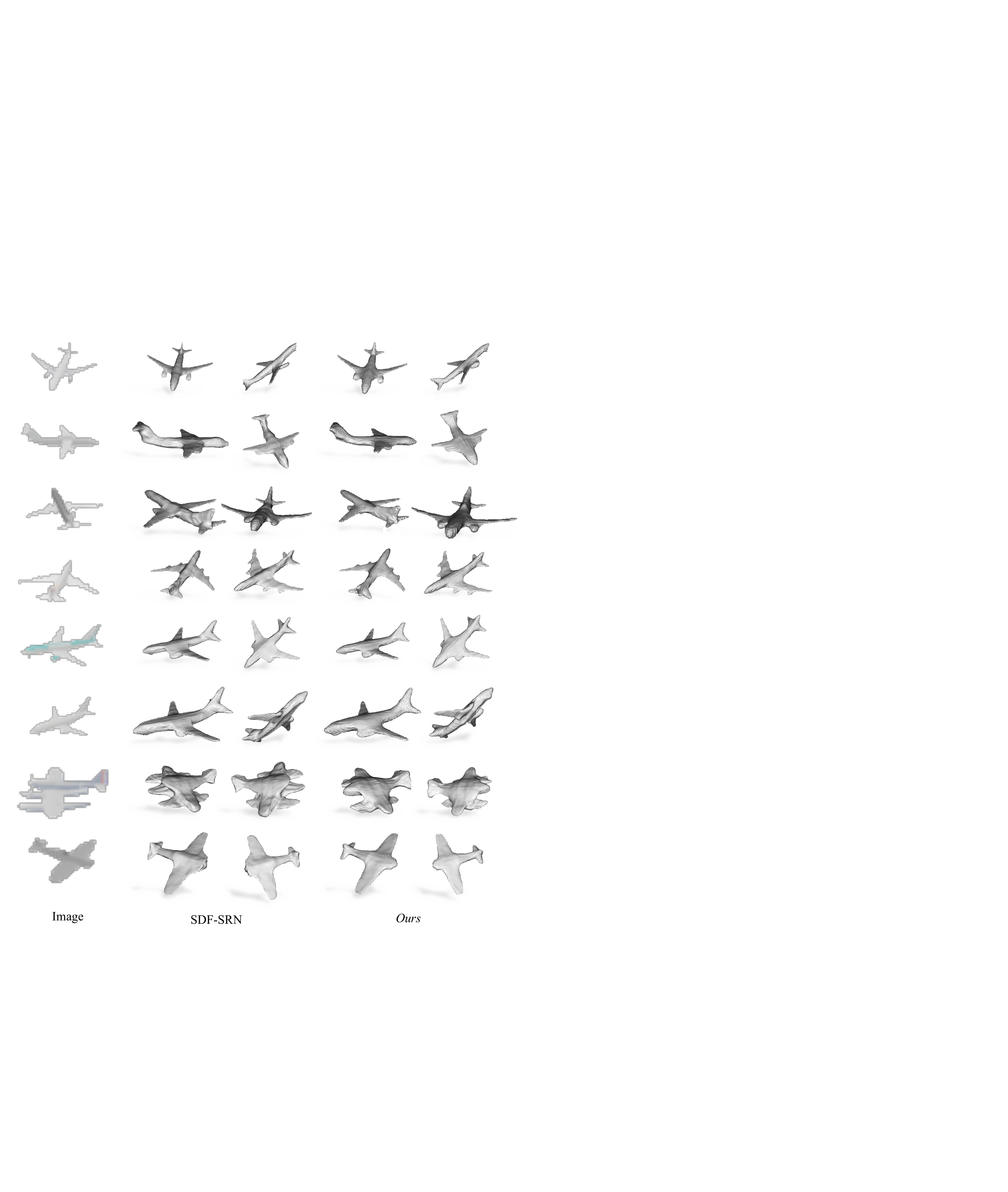}
\vspace{-2mm}
\caption{\textbf{3D Reconstruction on Shapenet airplanes from Single 2D Image}}
\label{fig:supp_shapenet_comparison_planes}
\vspace{-5mm}
\end{figure*}

\vspace{-2mm}
\paragraph{Single-View 3D Reconstruction on Pascal3D+ (unbiased) planes:} Figure~\ref{fig:supp_pascal_comparison_planes_unbiased} and Figure~\ref{fig:supp_pascal_comparison_planes_unbiased_v1} showcases additional results on the unbiased Pascal3D+ planes dataset. While, in comparison to the reconstructed planes of the default Pascal3D dataset, the unbiased Pascal3D reconstructions are of comparatively low fidelity, our approach still demonstrates significant improvement (in terms of less noisy reconstructions with better overall 3D structure) over the prior state of the art works of CMR \cite{cmrKanazawa18} (as shown in paper Figure 3) and SDF-SRN \cite{SDF-SRN}. The last two rows of Figure~\ref{fig:supp_pascal_comparison_planes_unbiased} and Figure~\ref{fig:supp_pascal_comparison_planes_unbiased_v1} showcase the examples where the input images (captured at the specific input viewpoints) do not provide enough geometric cues to the reconstruction pipeline to enable high-fidelity reconstruction. \emph{The significantly smaller size of the Pascal3D+ planes dataset is potentially the core reason behind such failures.}

\begin{figure*}
\centering
\includegraphics[width=0.70\linewidth]{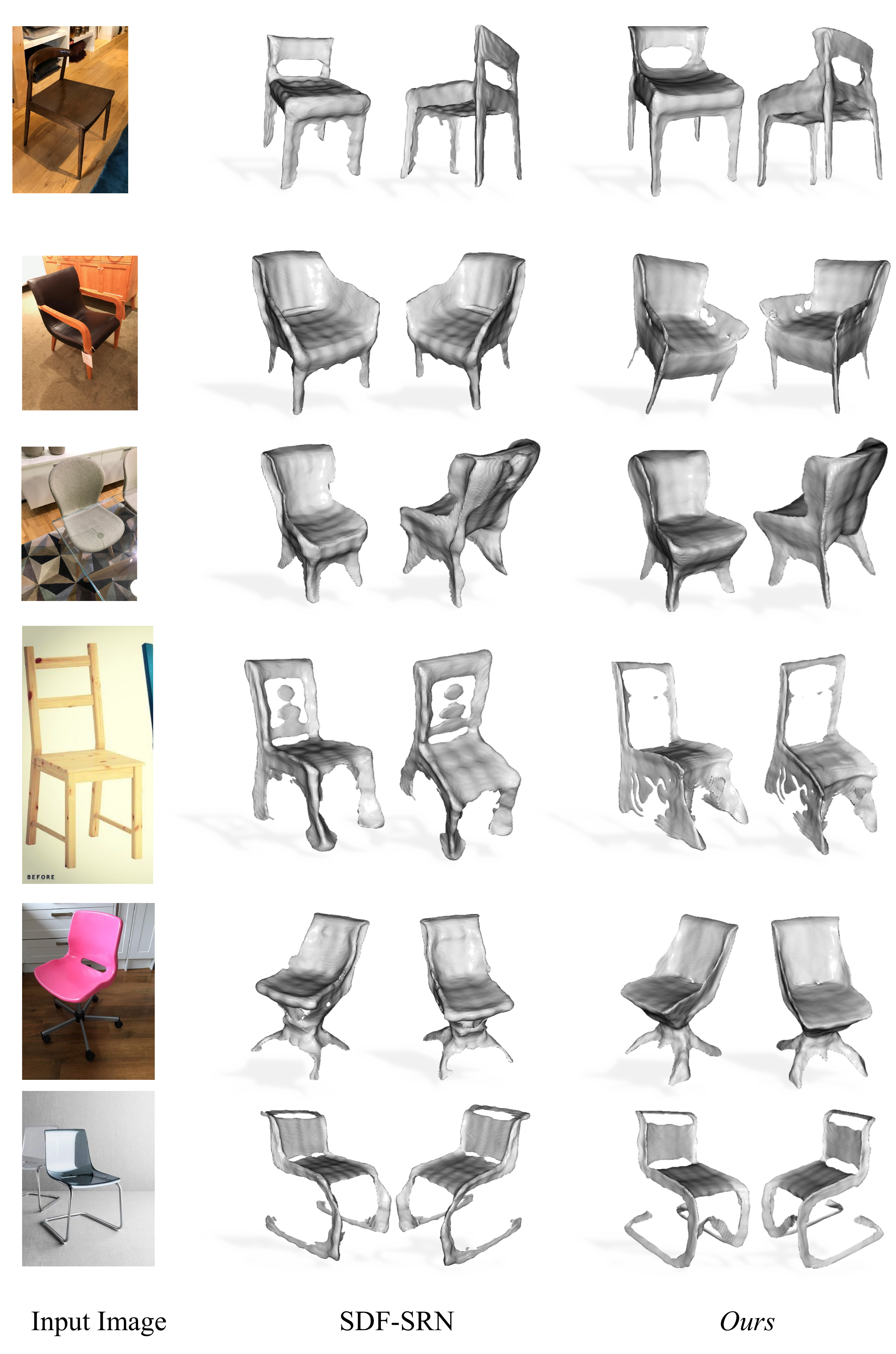}
\vspace{-2mm}
\caption{\textbf{3D Reconstruction on Pix3D Chairs from Single 2D Image (trained on Shapenet, tested on Pix3D val)}}
\label{fig:supp_pix3d_shapenet_comparison}
\vspace{-5mm}
\end{figure*}

\begin{figure*}
\vspace{-12mm}
\centering
\includegraphics[width=0.7\linewidth]{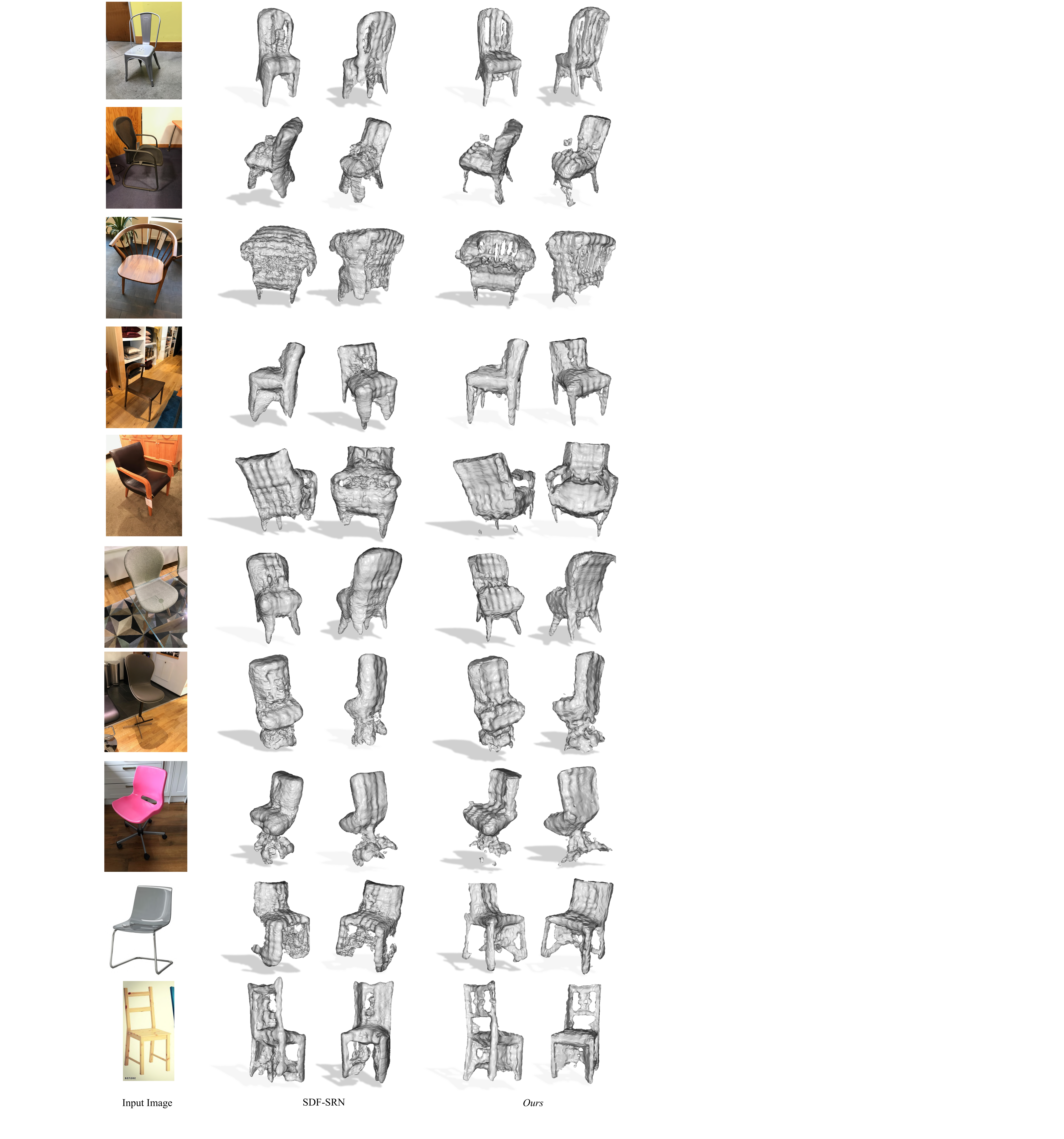}
\vspace{-2mm}
\caption{\textbf{3D Reconstruction on Pix3D Chairs (trained on Pix3D train + Pascal3D chairs, tested on Pix3D val)} }
\label{fig:supp_pix3d_comparison}
\vspace{-5mm}
\end{figure*}

\begin{figure*}
\centering
\includegraphics[width=0.85\linewidth]{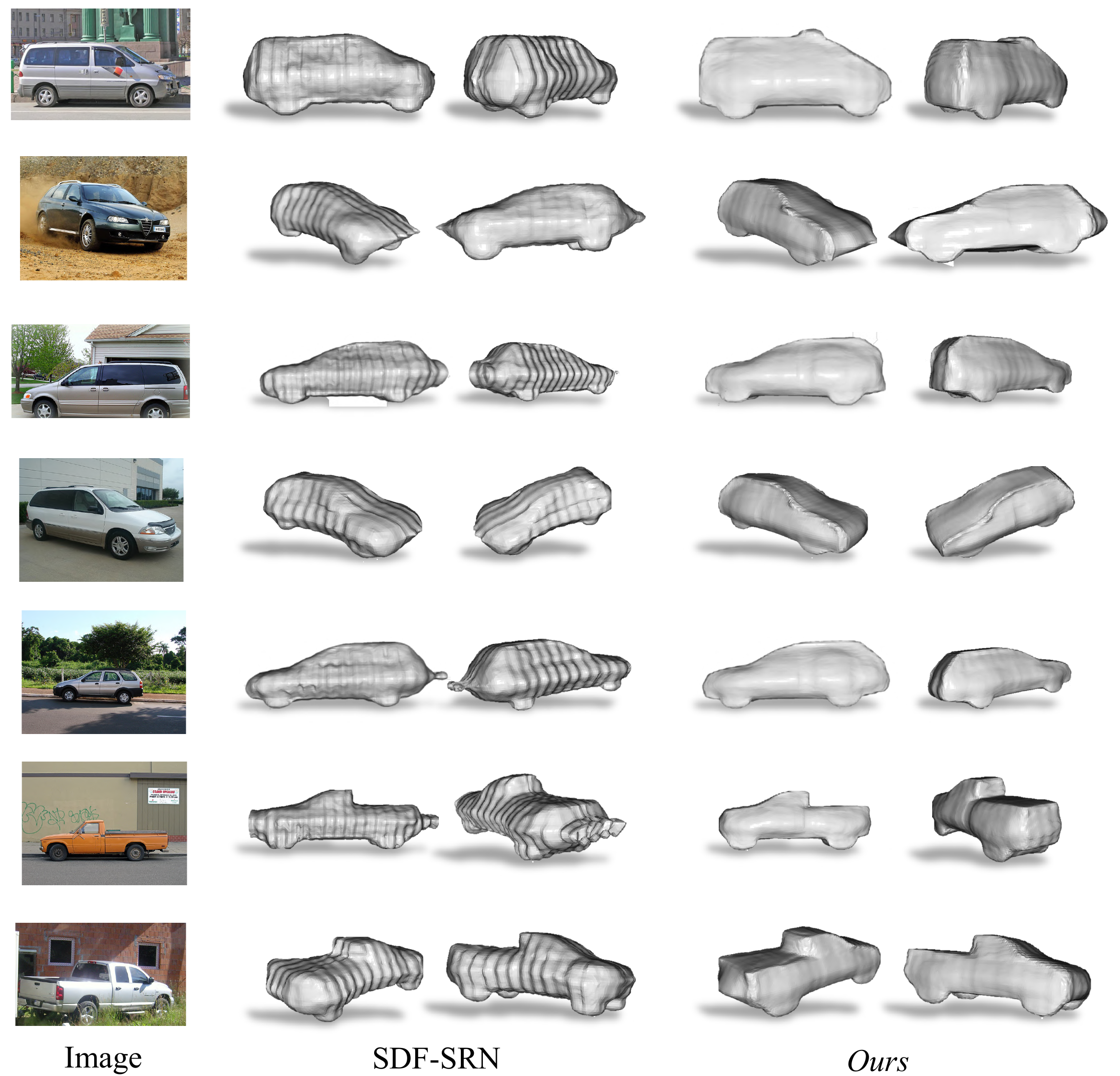}
\vspace{-2mm}
\caption{\textbf{3D Reconstruction on Pascal3D+ (default) Cars from Single 2D Image}}
\label{fig:supp_pascal_comparison_cars_default}
\vspace{-5mm}
\end{figure*}

\begin{figure*}
\centering
\includegraphics[width=0.85\linewidth]{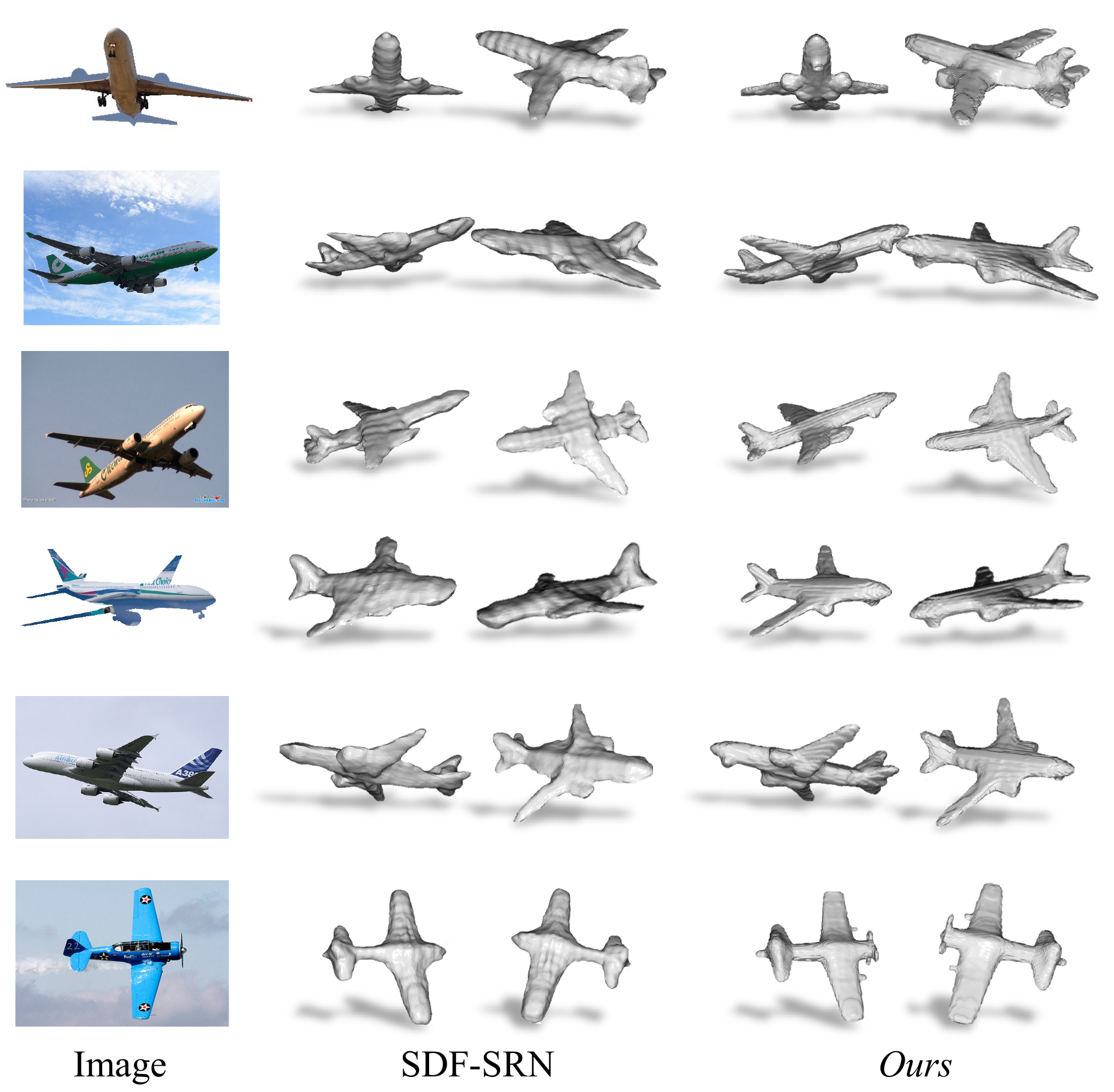}
\vspace{-2mm}
\caption{\textbf{3D Reconstruction on Pascal3D+ (default) Airplanes from Single 2D Image}}
\label{fig:supp_pascal_comparison_airplanes_default}
\vspace{-5mm}
\end{figure*}

\begin{figure*}
\centering
\includegraphics[width=0.99\linewidth]{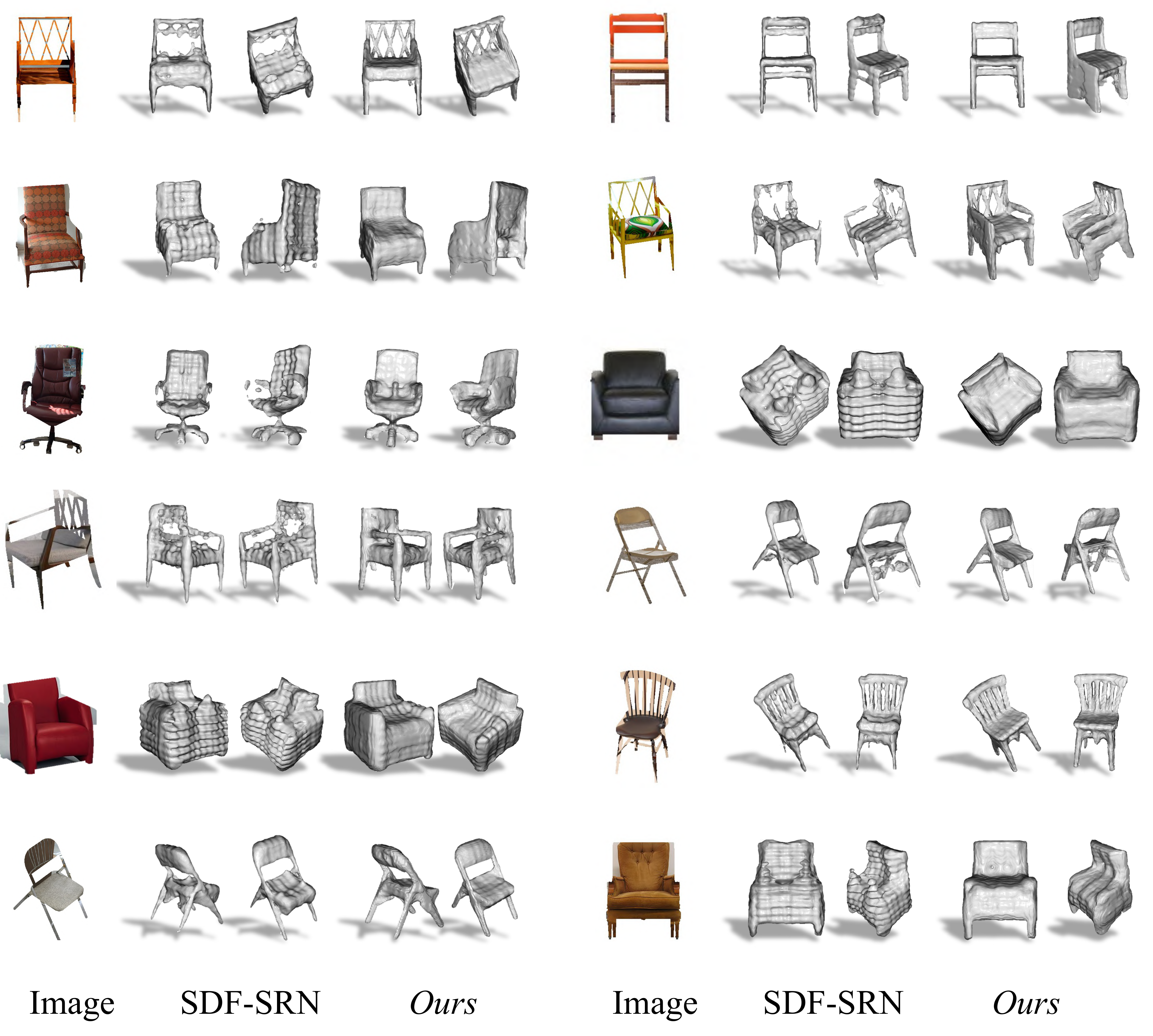}
\vspace{-2mm}
\caption{\textbf{3D Reconstruction on Pascal3D+ (default) Chairs from Single 2D Image:} Our approach not only yields high fidelity reconstructions, but also provides with dense cross-instance correspondences.}
\label{fig:supp_pascal_comparison_chairs_default}
\end{figure*}

\vspace{-3mm}
\paragraph{Single-View 3D Reconstruction on Pix3D Chairs:} We showcase additional qualitative comparison on the Pix3D chairs dataset in Figure~\ref{fig:supp_pix3d_shapenet_comparison} and Figure~\ref{fig:supp_pix3d_comparison}. Figure~\ref{fig:supp_pix3d_shapenet_comparison} highlights the synthetic to real generalization capability (trained on Shapenet, tested of Pix3D) of the reconstruction approaches. For Figure~\ref{fig:supp_pix3d_comparison}, we trained both our approach and SDF-SRN\cite{SDF-SRN} on the combined Pascal3D+ and Pix3D train dataset. While the results are much noisier (potentially because of smaller but challenging training set), the reconstructed shapes capture the overall geometry of the GT shapes and also maintain the topological structures of the GT chairs the majority of the times \emph{(see row 1, row 3, row 5 of Figure~\ref{fig:supp_pix3d_comparison})}. Like Pascal3D planes, the failure cases for the Pix3D chairs occur usually for input observations captured at some particular camera viewpoints which do not provide the reconstruction pipeline with enough geometric cues.

\begin{table}[t]
\centering
\scalebox{0.8}{
\begin{tabular}{llccccc}
\specialrule{.2em}{.1em}{.1em}
Method & \# training & \multicolumn{3}{c}{Chamfer $\downarrow$} \\
 & examples & acc. & cov. & overall \\
\specialrule{.1em}{.05em}{.05em}
\rowcolor{blue!7}
& 500 & \textbf{0.475} & 0.422 & \textbf{0.448}\\
\rowcolor{blue!7}
& 1000 & \textbf{0.442} & 0.385 & \textbf{0.413}\\
\rowcolor{blue!7} 
\multirow{-3}{*}{SDF-SRN \cite{SDF-SRN}} & 2000 & 0.423 & 0.349 & 0.386\\
\hline
\rowcolor{red!7}
& 500 & 0.495 & \textbf{0.402} & \textbf{0.448}\\
\rowcolor{red!7}
& 1000 & 0.462 & \textbf{0.366} & 0.414\\
\rowcolor{red!7} 
\multirow{-3}{*}{TARS (ours)} & 2000 & \textbf{0.423} & \textbf{0.347} & \textbf{0.385}\\
\specialrule{.1em}{.05em}{.05em}
\end{tabular}
}
\caption{\textbf{Dataset size ablation:} \# training examples vs reconstruction metrics.} 
\label{tab:dataset-ablation}
\end{table}

\begin{table}[t]
\centering
\scalebox{0.7}{
\begin{tabular}{llccccc}
\specialrule{.2em}{.1em}{.1em}
Method & Implicit & Dense & \multicolumn{3}{c}{Chamfer $\downarrow$} \\
& 3D Shape & Correspondences & acc. & cov. & overall \\
\specialrule{.1em}{.05em}{.05em}
SDF-SRN \cite{SDF-SRN} & \checkmark & & \textbf{0.352} & 0.315 & 0.333 \\
\hline
DIT \cite{zheng2020dit} & \checkmark & \checkmark & 0.386 & 0.326 & 0.356\\
DIF \cite{deng2021deformed} & \checkmark & \checkmark & 0.376 & 0.0308 & 0.342 \\
TARS (ours) & \checkmark & \checkmark & 0.353 & \textbf{0.312} & \textbf{0.332} \\
\specialrule{.1em}{.05em}{.05em}
\end{tabular}
}
\caption{\textbf{Comparison with Deformable Implicit Reconstruction approaches on ShapeNet Chairs dataset}} 
\label{tab:shapenet-comparison-deformable-implicit}
\vspace{-5mm}
\end{table}

\subsection{Quantitative Analysis}

\paragraph{Dataset size ablation:} We ablate the performance of our proposed approach as a factor of number of training examples on the Shapenet chairs dataset. For all the experiments under this ablation, we randomly sample a subset of CAD models. The training data is then generated by rendering each CAD model at only one randomly sampled viewpoint. From Table ~\ref{tab:dataset-ablation}, we see that both our proposed approach and SDF-SRN \cite{SDF-SRN} consistently performs well on all subsets of the Shapenet chairs dataset. Furthermore, increase in the dataset size does help the model achieve higher shape fidelity (in terms of reconstruction metrics). \sduggal{To re-emphasize on the need of larger training datasets: we think that comparatively less fidelity of the reconstructed real-world shapes (Pascal3D+, Pix3D) is because of the large variations (textural, environmental lightning, structural) in the real-world objects, but much smaller training datasets.}

\vspace{-3mm}
\paragraph{Comparison with Deformable Implicit Reconstruction approaches on Shapenet chairs dataset:} Recently, Zheng \etal \cite{zheng2020dit} and Deng \etal \cite{deng2021deformed} learned category-specific deformation fields and signed-distance fields jointly. While Zheng \etal \cite{zheng2020dit} \emph{(DIT: Deep Implicit Templates)} learned a 3D deformation field, Deng \etal \cite{deng2021deformed} \emph{(DIF: Deformed Implicit Fields)} learned a 3D deformation field + SDF correction field to handle the topological variations. Both the approaches required dense 3D supervision during training. \emph{In comparison to them, we address the task of single-view 3D reconstruction by learning higher-dimensional \emph{topologically-aware} deformation fields without using any form of dense supervision (multi-view images or dense 3D)}. In Table ~\ref{tab:shapenet-comparison-deformable-implicit}, we compare single-view analogs of their proposed approaches. We trained two ablations of our proposed approach: (a) single-view reconstruction using only 3D deformation fields (similar to the MLP-based deformation approach of Zheng \etal \cite{zheng2020dit}), (b) single-view 3D reconstruction using 3D deformation fields and 3D SDF correction fields (similar to Deng \etal \cite{deng2021deformed}). For both ablations, we do not learn any additional point features. We trained both the ablations using only single-view supervision exactly same as our proposed approach. From Table ~\ref{tab:shapenet-comparison-deformable-implicit}, we can see that solely learning deformation fields results in drop of the reconstruction metric (chamfer) compared to the reconstruction only approach (SDF-SRN). Adding SDF-correction fields on top of 3D deformation fields does ease the task of implicit deformation estimation and hence leads to better reconstructions. Our proposed approach performs better than both the deformation based implicit reconstruction approaches and is also at par with reconstruction only SDF-SRN approach \emph{(while learning deformations for free)}. 

\section{Future Work}

While, overall our results represent an encouraging step towards generalization of reconstruction systems to the internet of images, there is still more future work to be done to achieve scalable generalization. The immediate next step to unlock internet generalization is the removal of the requirement of known poses during training. Other directions could be the exploration of joint learning among multiple object categories, and efficient incorporation of adversarial learning to enable high-fidelity reconstruction even from input images captured at challenging viewpoints. Also, as we have witnessed the role of large annotated 2D datasets (like Imagenet \cite{Imagenet}) in rise of self-supervised learning in 2D \cite{moco, byol}, any potential work of generating much larger unbiased datasets like Pix3D could turn out to be a major step towards scalable single-view reconstruction.

\section{Statement on Potential Negative Impact}

We feel that the field of single-view 3D reconstruction is still in its nascent stage. So, we do not think this work has any immediate potential negative impact.


\vspace{-8mm}

%% file: main.bbl
\begin{thebibliography}{10}\itemsep=-1pt

\bibitem{QuickDraw}
https://quickdraw.withgoogle.com/.

\bibitem{achlioptas2017latent_pc}
Panos Achlioptas, Olga Diamanti, Ioannis Mitliagkas, and Leonidas~J Guibas.
\newblock Learning representations and generative models for 3d point clouds.
\newblock {\em arXiv preprint arXiv:1707.02392}, 2017.

\bibitem{nrsfm_kanade}
Ijaz Akhter, Yaser Sheikh, Sohaib Khan, and Takeo Kanade.
\newblock Nonrigid structure from motion in trajectory space.
\newblock In {\em NeurIPS}, 2008.

\bibitem{Bechtold2021HPN}
Jan Bechtold, Tatarchenko Maxim, Fischer Volker, and Brox Thomas.
\newblock Fostering generalization in single-view 3d reconstruction by learning
  a hierarchy of local and global shape priors.
\newblock In {\em CVPR}, 2021.

\bibitem{bhattad2021view}
Anand Bhattad, Aysegul Dundar, Guilin Liu, Andrew Tao, and Bryan Catanzaro.
\newblock View generalization for single image textured 3d models, 2021.

\bibitem{3DMM}
Volker Blanz and Thomas Vetter.
\newblock A morphable model for the synthesis of 3d faces.
\newblock In {\em Proceedings of the 26th Annual Conference on Computer
  Graphics and Interactive Techniques}, 1999.

\bibitem{Chang2015ShapeNetAI}
Angel~X. Chang, Thomas~A. Funkhouser, Leonidas~J. Guibas, Pat Hanrahan, Qixing
  Huang, Zimo Li, Silvio Savarese, Manolis Savva, Shuran Song, Hao Su,
  Jianxiong Xiao, L. Yi, and Fisher Yu.
\newblock Shapenet: An information-rich 3d model repository.
\newblock {\em ArXiv}, abs/1512.03012, 2015.

\bibitem{chen2019dibrender}
Wenzheng Chen, Jun Gao, Huan Ling, Edward Smith, Jaakko Lehtinen, Alec
  Jacobson, and Sanja Fidler.
\newblock Learning to predict 3d objects with an interpolation-based
  differentiable renderer.
\newblock In {\em NeurIPS}, 2019.

\bibitem{chen2018implicit_decoder}
Zhiqin Chen and Hao Zhang.
\newblock Learning implicit fields for generative shape modeling.
\newblock In {\em CVPR}, 2019.

\bibitem{SRF}
Julian Chibane, Aayush Bansal, Verica Lazova, and Gerard Pons-Moll.
\newblock Stereo radiance fields (srf): Learning view synthesis from sparse
  views of novel scenes.
\newblock In {\em CVPR}. {IEEE}, jun 2021.

\bibitem{choy20163d}
Christopher~B Choy, Danfei Xu, JunYoung Gwak, Kevin Chen, and Silvio Savarese.
\newblock 3d-r2n2: A unified approach for single and multi-view 3d object
  reconstruction.
\newblock In {\em ECCV}, 2016.

\bibitem{Imagenet}
Jia Deng, Wei Dong, Richard Socher, Li-Jia Li, Kai Li, and Li Fei-Fei.
\newblock Imagenet: A large-scale hierarchical image database.
\newblock In {\em CVPR}, 2009.

\bibitem{deng2021deformed}
Yu Deng, Jiaolong Yang, and Xin Tong.
\newblock Deformed implicit field: Modeling 3d shapes with learned dense
  correspondence.
\newblock In {\em CVPR}, 2021.

\bibitem{10.1007/BFb0086904}
A. Dervieux and F. Thomasset.
\newblock A finite element method for the simulation of a rayleigh-taylor
  instability.
\newblock In Reimund Rautmann, editor, {\em Approximation Methods for
  Navier-Stokes Problems}, 1980.

\bibitem{dosovitskiy2020vit}
Alexey Dosovitskiy, Lucas Beyer, Alexander Kolesnikov, Dirk Weissenborn,
  Xiaohua Zhai, Thomas Unterthiner, Mostafa Dehghani, Matthias Minderer, Georg
  Heigold, Sylvain Gelly, Jakob Uszkoreit, and Neil Houlsby.
\newblock An image is worth 16x16 words: Transformers for image recognition at
  scale.
\newblock {\em ICLR}, 2021.

\bibitem{sdf_secrets}
Shivam Duggal, Zihao Wang, Wei-Chiu Ma, Sivabalan Manivasagam, Justin Liang,
  Shenlong Wang, and Raquel Urtasun.
\newblock Mending neural implicit modeling for 3d vehicle reconstruction in the
  wild.
\newblock In {\em WACV}, 2022.

\bibitem{Engelmann2017SAMPSA}
Francis Engelmann, J. St{\"u}ckler, and B. Leibe.
\newblock Samp: Shape and motion priors for 4d vehicle reconstruction.
\newblock {\em WACV}, 2017.

\bibitem{PascalVOC}
Mark Everingham, Luc Gool, Christopher~K. Williams, John Winn, and Andrew
  Zisserman.
\newblock The pascal visual object classes (voc) challenge.
\newblock {\em Int. J. Comput. Vision}, 88(2):303–338, jun 2010.

\bibitem{ucmrGoel20}
Shubham Goel, Angjoo Kanazawa, , and Jitendra Malik.
\newblock Shape and viewpoints without keypoints.
\newblock In {\em ECCV}, 2020.

\bibitem{byol}
Jean{-}Bastien Grill, Florian Strub, Florent Altch{\'{e}}, Corentin Tallec,
  Pierre~H. Richemond, Elena Buchatskaya, Carl Doersch, Bernardo~{\'{A}}vila
  Pires, Zhaohan~Daniel Guo, Mohammad~Gheshlaghi Azar, Bilal Piot, Koray
  Kavukcuoglu, R{\'{e}}mi Munos, and Michal Valko.
\newblock Bootstrap your own latent: {A} new approach to self-supervised
  learning.
\newblock {\em CoRR}, abs/2006.07733, 2020.

\bibitem{moco}
Kaiming He, Haoqi Fan, Yuxin Wu, Saining Xie, and Ross~B. Girshick.
\newblock Momentum contrast for unsupervised visual representation learning.
\newblock {\em CoRR}, abs/1911.05722, 2019.

\bibitem{mask-rcnn}
Kaiming He, Georgia Gkioxari, Piotr Dollár, and Ross Girshick.
\newblock Mask r-cnn.
\newblock In {\em ICCV}, 2017.

\bibitem{He2015}
Kaiming He, Xiangyu Zhang, Shaoqing Ren, and Jian Sun.
\newblock Deep residual learning for image recognition.
\newblock {\em arXiv preprint arXiv:1512.03385}, 2015.

\bibitem{homayounfar2020levelset}
Namdar Homayounfar, Yuwen Xiong, Justin Liang, Wei-Chiu Ma, and Raquel Urtasun.
\newblock Levelset r-cnn: A deep variational method for instance segmentation,
  2020.

\bibitem{qixing_huang_retrieval}
Qixing Huang, Hai Wang, and Vladlen Koltun.
\newblock Single-view reconstruction via joint analysis of image and shape
  collections.
\newblock {\em ACM Trans. Graph.}, 34(4), July 2015.

\bibitem{jain2021putting}
Ajay Jain, Matthew Tancik, and Pieter Abbeel.
\newblock Putting nerf on a diet: Semantically consistent few-shot view
  synthesis, 2021.

\bibitem{cmrKanazawa18}
Angjoo Kanazawa, Shubham Tulsiani, Alexei~A. Efros, and Jitendra Malik.
\newblock Learning category-specific mesh reconstruction from image
  collections.
\newblock In {\em ECCV}, 2018.

\bibitem{Kar2015CategoryspecificOR}
Abhishek Kar, Shubham Tulsiani, Jo{\~a}o Carreira, and Jitendra Malik.
\newblock Category-specific object reconstruction from a single image.
\newblock {\em CVPR}, pages 1966--1974, 2015.

\bibitem{kato2018renderer}
Hiroharu Kato, Yoshitaka Ushiku, and Tatsuya Harada.
\newblock Neural 3d mesh renderer.
\newblock In {\em CVPR}, 2018.

\bibitem{shape_from_silhouette}
Aldo Laurentini.
\newblock The visual hull concept for silhouette-based image understanding.
\newblock {\em Pattern Analysis and Machine Intelligence, IEEE Transactions
  on}, 16:150--162, 03 1994.

\bibitem{LiHM15}
Ke Li, Bharath Hariharan, and Jitendra Malik.
\newblock Iterative instance segmentation.
\newblock {\em CoRR}, abs/1511.08498, 2015.

\bibitem{umr2020}
Xueting Li, Sifei Liu, Kihwan Kim, Shalini De~Mello, Varun Jampani, Ming-Hsuan
  Yang, and Jan Kautz.
\newblock Self-supervised single-view 3d reconstruction via semantic
  consistency.
\newblock In {\em ECCV}, 2020.

\bibitem{SDF-SRN}
Chen-Hsuan Lin, Chaoyang Wang, and Simon Lucey.
\newblock Sdf-srn: Learning signed distance 3d object reconstruction from
  static images.
\newblock In {\em NeurIPS}, 2020.

\bibitem{liu2019morphing}
Minghua Liu, Lu Sheng, Sheng Yang, Jing Shao, and Shi-Min Hu.
\newblock Morphing and sampling network for dense point cloud completion.
\newblock {\em arXiv preprint arXiv:1912.00280}, 2019.

\bibitem{liu2019softras}
Shichen Liu, Tianye Li, Weikai Chen, and Hao Li.
\newblock Soft rasterizer: A differentiable renderer for image-based 3d
  reasoning.
\newblock {\em ICCV}, Oct 2019.

\bibitem{dist}
Shaohui Liu, Yinda Zhang, Songyou Peng, Boxin Shi, Marc Pollefeys, and Zhaopeng
  Cui.
\newblock Dist: Rendering deep implicit signed distance function with
  differentiable sphere tracing.
\newblock In {\em CVPR}, 2020.

\bibitem{liu2020dist}
Shaohui Liu, Yinda Zhang, Songyou Peng, Boxin Shi, Marc Pollefeys, and Zhaopeng
  Cui.
\newblock Dist: Rendering deep implicit signed distance function with
  differentiable sphere tracing.
\newblock In {\em CVPR}, 2020.

\bibitem{Loper2014OpenDRAA}
Matthew Loper and Michael~J. Black.
\newblock Opendr: An approximate differentiable renderer.
\newblock In {\em ECCV}, 2014.

\bibitem{SMPL:2015}
Matthew Loper, Naureen Mahmood, Javier Romero, Gerard Pons-Moll, and Michael~J.
  Black.
\newblock {SMPL}: A skinned multi-person linear model.
\newblock {\em ACM Trans. Graphics (Proc. SIGGRAPH Asia)}, 34(6):248:1--248:16,
  Oct. 2015.

\bibitem{SMPL}
Matthew Loper, Naureen Mahmood, Javier Romero, Gerard Pons-Moll, and Michael~J.
  Black.
\newblock {SMPL}: A skinned multi-person linear model.
\newblock {\em ACM Trans. Graphics (Proc. SIGGRAPH Asia)}, 34(6):248:1--248:16,
  Oct. 2015.

\bibitem{MarchingCubes}
William~E. Lorensen and Harvey~E. Cline.
\newblock Marching cubes: A high resolution 3d surface construction algorithm.
\newblock In {\em SIGGRAPH}, 1987.

\bibitem{Luo-VideoDepth-2020}
Xuan Luo, Jia-Bin Huang, Richard Szeliski, Kevin Matzen, and Johannes Kopf.
\newblock Consistent video depth estimation.
\newblock {\em ACM Transactions on Graphics (TOG)}, 39(4):71--1, 2020.

\bibitem{ONet}
Lars~M. Mescheder, Michael Oechsle, M. Niemeyer, Sebastian Nowozin, and Andreas
  Geiger.
\newblock Occupancy networks: Learning 3d reconstruction in function space.
\newblock In {\em CVPR}, 2019.

\bibitem{mildenhall2020nerf}
Ben Mildenhall, Pratul~P. Srinivasan, Matthew Tancik, Jonathan~T. Barron, Ravi
  Ramamoorthi, and Ren Ng.
\newblock Nerf: Representing scenes as neural radiance fields for view
  synthesis.
\newblock In {\em ECCV}, 2020.

\bibitem{Newcombe2011KinectFusionRD}
Richard~A. Newcombe, Shahram Izadi, Otmar Hilliges, David Molyneaux, David Kim,
  Andrew~J. Davison, Pushmeet Kohli, Jamie Shotton, Steve Hodges, and Andrew~W.
  Fitzgibbon.
\newblock Kinectfusion: Real-time dense surface mapping and tracking.
\newblock {\em 2011 10th IEEE International Symposium on Mixed and Augmented
  Reality}, pages 127--136, 2011.

\bibitem{point_set_distance}
Trung Nguyen, Quang-Hieu Pham, Tam Le, Tung Pham, Nhat Ho, and Binh-Son Hua.
\newblock Point-set distances for learning representations of 3d point clouds.
\newblock {\em arXiv preprint arXiv:2102.04014}, 2021.

\bibitem{niemeyer2020differentiable}
Michael Niemeyer, Lars Mescheder, Michael Oechsle, and Andreas Geiger.
\newblock Differentiable volumetric rendering: Learning implicit 3d
  representations without 3d supervision.
\newblock In {\em CVPR}, 2020.

\bibitem{DVR}
Michael Niemeyer, Lars Mescheder, Michael Oechsle, and Andreas Geiger.
\newblock Differentiable volumetric rendering: Learning implicit 3d
  representations without 3d supervision.
\newblock In {\em CVPR}, 2020.

\bibitem{Oechsle2021ICCV}
Michael Oechsle, Songyou Peng, and Andreas Geiger.
\newblock Unisurf: Unifying neural implicit surfaces and radiance fields for
  multi-view reconstruction.
\newblock In {\em ICCV}, 2021.

\bibitem{Osher01levelset}
Stanley Osher and Ronald~P. Fedkiw.
\newblock Level set methods: An overview and some recent results.
\newblock {\em J. Comput. Phys}, 169:463--502, 2001.

\bibitem{Osher88frontspropagating}
Stanley Osher and James~A. Sethian.
\newblock Fronts propagating with curvature dependent speed: algorithms based
  on hamilton–jacobi formulations.
\newblock {\em Journal of Computational Physics}, pages 12--49, 1988.

\bibitem{Ovsjanikov_functionalmaps:}
Maks Ovsjanikov, Mirela Ben-chen, Justin Solomon, Adrian Butscher, Leonidas
  Guibas, and Lix École Polytechnique.
\newblock Functional maps: A flexible representation of maps between shapes.

\bibitem{park2019deepsdf}
Jeong~Joon Park, Peter Florence, Julian Straub, Richard Newcombe, and Steven
  Lovegrove.
\newblock Deepsdf: Learning continuous signed distance functions for shape
  representation.
\newblock In {\em CVPR}, 2019.

\bibitem{park2021nerfies}
Keunhong Park, Utkarsh Sinha, Jonathan~T. Barron, Sofien Bouaziz, Dan~B
  Goldman, Steven~M. Seitz, and Ricardo Martin-Brualla.
\newblock Nerfies: Deformable neural radiance fields.
\newblock {\em ICCV}, 2021.

\bibitem{park2021hypernerf}
Keunhong Park, Utkarsh Sinha, Peter Hedman, Jonathan~T. Barron, Sofien Bouaziz,
  Dan~B Goldman, Ricardo Martin-Brualla, and Steven~M. Seitz.
\newblock Hypernerf: A higher-dimensional representation for topologically
  varying neural radiance fields.
\newblock {\em arXiv preprint arXiv:2106.13228}, 2021.

\bibitem{PiFU}
Shunsuke Saito, , Zeng Huang, Ryota Natsume, Shigeo Morishima, Angjoo Kanazawa,
  and Hao Li.
\newblock Pifu: Pixel-aligned implicit function for high-resolution clothed
  human digitization.
\newblock In {\em ICCV}, 2019.

\bibitem{schoenberger2016mvs}
Johannes~Lutz Sch\"{o}nberger, Enliang Zheng, Marc Pollefeys, and Jan-Michael
  Frahm.
\newblock {Pixelwise View Selection for Unstructured Multi-View Stereo}.
\newblock In {\em ECCV}, 2016.

\bibitem{MetaSDF}
Vincent Sitzmann, Eric~R. Chan, Richard Tucker, Noah Snavely, and Gordon
  Wetzstein.
\newblock Metasdf: Meta-learning signed distance functions.
\newblock In {\em NeurIPS}, 2020.

\bibitem{sitzmann2019siren}
Vincent Sitzmann, Julien~N.P. Martel, Alexander~W. Bergman, David~B. Lindell,
  and Gordon Wetzstein.
\newblock Implicit neural representations with periodic activation functions.
\newblock In {\em arXiv}, 2020.

\bibitem{SRN}
Vincent Sitzmann, Michael Zollh{\"o}fer, and Gordon Wetzstein.
\newblock Scene representation networks: Continuous 3d-structure-aware neural
  scene representations.
\newblock In {\em NeurIPS}, 2019.

\bibitem{pix3d}
Xingyuan Sun, Jiajun Wu, Xiuming Zhang, Zhoutong Zhang, Chengkai Zhang, Tianfan
  Xue, Joshua~B Tenenbaum, and William~T Freeman.
\newblock Pix3d: Dataset and methods for single-image 3d shape modeling.
\newblock In {\em CVPR}, 2018.

\bibitem{Tatarchenko2019WhatDS}
Maxim Tatarchenko, Stephan~R. Richter, Ren{\'e} Ranftl, Zhuwen Li, Vladlen
  Koltun, and Thomas Brox.
\newblock What do single-view 3d reconstruction networks learn?
\newblock {\em CVPR}, 2019.

\bibitem{Tewari2020NeuralSTAR}
A. Tewari, O. Fried, J. Thies, V. Sitzmann, S. Lombardi, K. Sunkavalli, R.
  Martin-Brualla, T. Simon, J. Saragih, M. Nie{\ss}ner, R. Pandey, S. Fanello,
  G. Wetzstein, J.-Y. Zhu, C. Theobalt, M. Agrawala, E. Shechtman, D.~B
  Goldman, and M. Zollh{\"o}fer.
\newblock {State of the Art on Neural Rendering}.
\newblock {\em Computer Graphics Forum (EG STAR 2020)}, 2020.

\bibitem{neural-rendering}
Ayush Tewari, Ohad Fried, Justus Thies, Vincent Sitzmann, Stephen Lombardi,
  Kalyan Sunkavalli, Ricardo Martin{-}Brualla, Tomas Simon, Jason~M. Saragih,
  Matthias Nie{\ss}ner, Rohit Pandey, Sean~Ryan Fanello, Gordon Wetzstein,
  Jun{-}Yan Zhu, Christian Theobalt, Maneesh Agrawala, Eli Shechtman, Dan~B.
  Goldman, and Michael Zollh{\"{o}}fer.
\newblock State of the art on neural rendering.
\newblock {\em CoRR}, abs/2004.03805, 2020.

\bibitem{Face2Face}
Justus Thies, Michael Zollh{\"{o}}fer, Marc Stamminger, Christian Theobalt, and
  Matthias Nie{\ss}ner.
\newblock Face2face: Real-time face capture and reenactment of {RGB} videos.
\newblock {\em CoRR}, abs/2007.14808, 2020.

\bibitem{tretschk2021nonrigid}
Edgar Tretschk, Ayush Tewari, Vladislav Golyanik, Michael Zollh\"{o}fer,
  Christoph Lassner, and Christian Theobalt.
\newblock Non-rigid neural radiance fields: Reconstruction and novel view
  synthesis of a dynamic scene from monocular video.
\newblock In {\em ICCV}. {IEEE}, 2021.

\bibitem{implciit_mesh_shubham}
Shubham Tulsiani, Nilesh Kulkarni, and Abhinav Gupta.
\newblock Implicit mesh reconstruction from unannotated image collections.
\newblock {\em CoRR}, abs/2007.08504, 2020.

\bibitem{drcTulsiani17}
Shubham Tulsiani, Tinghui Zhou, Alexei~A. Efros, and Jitendra Malik.
\newblock Multi-view supervision for single-view reconstruction via
  differentiable ray consistency.
\newblock In {\em CVPR}, 2017.

\bibitem{TulsianiZEM17}
Shubham Tulsiani, Tinghui Zhou, Alexei~A. Efros, and Jitendra Malik.
\newblock Multi-view supervision for single-view reconstruction via
  differentiable ray consistency.
\newblock {\em CoRR}, abs/1704.06254, 2017.

\bibitem{wang2020deep}
Chaoyang Wang, Chen-Hsuan Lin, and Simon Lucey.
\newblock Deep nrsfm++: Towards unsupervised 2d-3d lifting in the wild.
\newblock In {\em 3DV}, 2020.

\bibitem{wang2021neus}
Peng Wang, Lingjie Liu, Yuan Liu, Christian Theobalt, Taku Komura, and Wenping
  Wang.
\newblock Neus: Learning neural implicit surfaces by volume rendering for
  multi-view reconstruction.
\newblock {\em arXiv preprint arXiv:2106.10689}, 2021.

\bibitem{Wang2019DirectShapePA}
Rui Wang, Nan Yang, J. St{\"u}ckler, and Daniel Cremers.
\newblock Directshape: Photometric alignment of shape priors for visual vehicle
  pose and shape estimation.
\newblock {\em ArXiv}, abs/1904.10097, 2019.

\bibitem{WelinderEtal2010}
P. Welinder, S. Branson, T. Mita, C. Wah, F. Schroff, S. Belongie, and P.
  Perona.
\newblock {Caltech-UCSD Birds 200}.
\newblock Technical report, California Institute of Technology, 2010.

\bibitem{weng2018photo}
Chung-Yi Weng, Brian Curless, and Ira Kemelmacher-Shlizerman.
\newblock Photo wake-up: 3d character animation from a single photo, 2018.

\bibitem{marrnet}
Jiajun Wu, Yifan Wang, Tianfan Xue, Xingyuan Sun, William~T Freeman, and
  Joshua~B Tenenbaum.
\newblock {MarrNet: 3D Shape Reconstruction via 2.5D Sketches}.
\newblock In {\em NeurIPS}, 2017.

\bibitem{shapehd}
Jiajun Wu, Chengkai Zhang, Xiuming Zhang, Zhoutong Zhang, William~T Freeman,
  and Joshua~B Tenenbaum.
\newblock {Learning 3D Shape Priors for Shape Completion and Reconstruction}.
\newblock In {\em ECCV}, 2018.

\bibitem{Pascal3D}
Yu Xiang, Roozbeh Mottaghi, and Silvio Savarese.
\newblock Beyond pascal: A benchmark for 3d object detection in the wild.
\newblock In {\em WACV}, 2014.

\bibitem{yariv2020multiview}
Lior Yariv, Yoni Kasten, Dror Moran, Meirav Galun, Matan Atzmon, Basri Ronen,
  and Yaron Lipman.
\newblock Multiview neural surface reconstruction by disentangling geometry and
  appearance.
\newblock {\em NeurIPS}, 33, 2020.

\bibitem{yu2020pixelnerf}
Alex Yu, Vickie Ye, Matthew Tancik, and Angjoo Kanazawa.
\newblock pixelnerf: Neural radiance fields from one or few images, 2020.

\bibitem{sdflabel}
Sergey Zakharov, Wadim Kehl, Arjun Bhargava, and Adrien Gaidon.
\newblock Autolabeling 3d objects with differentiable rendering of sdf shape
  priors.
\newblock In {\em CVPR}, June 2020.

\bibitem{zbontar2016stereo}
Jure Zbontar and Yann LeCun.
\newblock Stereo matching by training a convolutional neural network to compare
  image patches.
\newblock {\em Journal of Machine Learning Research}, 17:1--32, 2016.

\bibitem{zhang2021ners}
Jason~Y. Zhang, Gengshan Yang, Shubham Tulsiani, and Deva Ramanan.
\newblock {NeRS}: Neural reflectance surfaces for sparse-view 3d reconstruction
  in the wild.
\newblock In {\em Conference on Neural Information Processing Systems}, 2021.

\bibitem{zheng2020dit}
Zerong Zheng, Tao Yu, Qionghai Dai, and Yebin Liu.
\newblock Deep implicit templates for 3d shape representation, 2020.

\bibitem{SMAL}
Silvia Zuffi, Angjoo Kanazawa, David Jacobs, and Michael~J. Black.
\newblock {3D} menagerie: Modeling the {3D} shape and pose of animals.
\newblock In {\em CVPR}, July 2017.

\end{thebibliography}
